\documentclass[12pt,A4]{article}
\usepackage[margin=1in,footskip=0.25in]{geometry}

\usepackage{graphicx, color}  
\usepackage{bm}
\usepackage{amsmath, amsfonts, MnSymbol, mathrsfs}

\usepackage{amsmath}
\usepackage[noend]{algpseudocode}
\usepackage{algorithmicx,algorithm}
\usepackage{multirow}
\usepackage{booktabs}
\usepackage{threeparttable}
\usepackage{hyperref}
\usepackage{enumerate}
\usepackage{multirow}
\usepackage{subfig}

\hypersetup{
	colorlinks=true,
	linkcolor=black,
	filecolor=black,      
	urlcolor=black,
	citecolor=black,
}

\def\cn{\centerline}

\def\aletheia{{\bf Aletheia}$^\copyright$ }

 \def\C{\bm{C}}

\def\b{\bm{b}} \def\x{\bm{x}} \def\y{\bm{y}}  \def\z{\bm{z}}
\def\u{\bm{u}}  \def\w{\bm{w}} 
\def\D{\bm{D}}  \def\X{\bm{X}} 
\def\P{\bm{P}}  \def\W{\bm{W}}

 \def \bb{\bm\beta} 
 \def\thth{\bm\theta}

\def\wtilde{\bm{\tilde{w}}}

\def \NN {\mathcal{N}}   \def \CC {\mathcal{C}}
\def \RR {\mathcal{R}} \def \XX {\mathcal{X}}
\def \LL {\mathcal{L}}  \def \PP {\mathscr{P}}

\def \RRR {\mathbb{R}} 

\def \EEE {\mathbb{E}}

\def \ve{\varepsilon}

 \def \zero {\mathbf{0}}

\newtheorem{lemma}{Lemma}
\newtheorem{theorem}{Theorem}

\newtheorem{definition}{Definition}

\begin{document}
	
	\title{Unwrapping The Black Box of Deep ReLU Networks: Interpretability, Diagnostics, and Simplification} 

	\def\thefootnote{*}\footnotetext{Corresponding author: ajzhang@umich.edu}
	\def\thefootnote{\arabic{footnote}}	
	
	\author{Agus Sudjianto$^1$, William Knauth$^2$,  Rahul Singh$^1$\\
		Zebin Yang$^3$, and Aijun Zhang$^{3,*}$}
	\date{\normalsize $^1$Wells Fargo, 
		$^2$Carnegie Mellon University, 
		$^3$The University of Hong Kong} 
	\maketitle

	\begin{abstract} 
		The deep neural networks (DNNs) have achieved great success in learning complex patterns with strong predictive power, but they are often thought of as ``black box'' models without a sufficient level of transparency and interpretability. It is important to demystify the DNNs with rigorous mathematics and practical tools, especially when they are used for mission-critical applications. This paper aims to unwrap the black box of deep ReLU networks through local linear representation, which utilizes the activation pattern and disentangles the complex network into an equivalent set of local linear models (LLMs). We develop a convenient LLM-based toolkit for interpretability, diagnostics, and simplification of a pre-trained deep ReLU network. We propose the local linear profile plot and other visualization methods for interpretation and diagnostics, and an effective merging strategy for network simplification. 
The proposed methods are demonstrated by simulation examples, benchmark datasets, and a real case study in home lending credit risk assessment.
		
		
		\vskip 6.5pt \noindent {\bf Keywords}: Activation pattern, Deep ReLU network, Local linear interpretability, Network diagnostics, Network simplification.
	\end{abstract}

	\section{Introduction}
	Explainable artificial intelligence (XAI) or interpretable machine learning (IML) has received great attention in recent years; see \cite{gunning2019darpa, molnar2020interpretable, murdoch2019definitions, arrieta2020explainable} among others. The majority of machine learning algorithms have the ability to accurately predict various complex phenomena,  but they are often regarded as black-box models since their inner decision-making process cannot be easily understood by human beings.  A fundamental objective of XAI/IML is to understand why these black-box models work for different real-world problems.  This is important for AI/ML model usage in mission-critical applications, not only due to the increased regulatory scrutiny but also in order to gain the trust of the model users. Without a good understanding of why an AI/ML model works,  it may perform abnormally as the data feed is slightly changed, for an instance, the adversarial perturbation in computer vision \cite{goodfellow2014explaining}.
	
	There exist two lines of XAI/IML research works about model interpretation, namely intrinsic model interpretability and post-hoc model explainability. Intrinsic interpretability is to study the inherent statistical properties of a specific machine learning model. It is argued by \cite{yang2020enhancing} that the intrinsic interpretability of a complex model ought to be induced from practical constraints, including additivity, sparsity, monotonicity, near-orthogonality, linearity, and smoothness. In this regard, some classical models in statistics are actually intrinsically interpretable, including generalized linear model (GLM) \cite{mccullagh1989generalized}, generalized additive model (GAM) \cite{hastie1990generalized}, and generalized additive index model (GAIM, also known as projection pursuit learning) \cite{friedman1981projection, hwang1994regression}. 
	In the new IML literature, intrinsically interpretable models that integrate deep neural networks include GAM-based  NAM \cite{agarwal2020neural} 
	and GAMI-Net \cite{yang2020gami}, and GAIM-based   xNN \cite{vaughan2018explainable}, ExNN \cite{yang2020enhancing} and AxNN \cite{chen2020adaptive}.  In computer vision, there  exist intrinsically interpretable models based on convolutional neural networks \cite{zhang2018interpretable, angelov2020towards, bau2020understanding}. 
	
	On the other hand, post-hoc explainability is model-agnostic. 
	For any black-box model that produces an output from an input,  post-hoc analytical methods include variable importance (VI) \cite{breiman2001random}, partial dependence plot (PDP) \cite{friedman2001greedy},  individual conditional expectation (ICE) plot \cite{goldstein2015peeking}, and accumulated local effects (ALE) plot \cite{apley2020visualizing}. Post-hoc explainability may also rely on surrogate modeling, including both the global surrogates as in model compression or distillation \cite{bucilua2006model,ba2014deep, hinton2015distilling} and the perturbation-based local surrogates as in LIME \cite{ribeiro2016should} and SHAP \cite{lundberg2017unified, lundberg2020local}. However, these post-hoc methods are often based on crude assumptions, and their interpretation results are not fully reliable. For an instance, \cite{slack2020fooling} recently proposed an adversarial attack method that may hide the model bias from LIME and SHAP.
	
	In this paper, we investigate the intrinsic interpretability of deep neural networks (DNNs) with ReLU (rectified linear unit) activation functions, which are arguably the most popular type of neural networks in deep learning \cite{lecun2015deep, goodfellow2016deep}. The deep ReLU networks enjoy many appealing properties, including better performance than sigmoidal activations \cite{glorot2011deep},  rich expressivity, and universal approximation ability \cite{montufar2014number, lu2017expressive, arora2018understanding, schmidt2020nonparametric}, double-descent risk curve upon over-parameterization \cite{belkin2019reconciling}, fast and scalable training algorithms by stochastic gradient descent (SGD) and its variants \cite{zou2020gradient}.  Despite their strong predictive power, the deep ReLU networks act as a black box without a sufficient level of transparency, interpretability, and robustness. The related works on ReLU DNN interpretation are reviewed below.
	
	\subsection{Related Works}
	Besides the model-agnostic post-hoc methods reviewed above, there exist some works that tried to interpret pre-trained deep ReLU networks specifically. One type of existing works is to quantify the individual feature importance by computing integrated gradients \cite{sundararajan2017axiomatic} or decomposing predictions through layer-wise backpropagation \cite{bach2015pixel, shrikumar2017learning}, as an attempt to attribute the influence on prediction to each input feature in a rank-order manner. There exist similar methods for detecting feature interactions in deep ReLU networks,  including pairwise feature interactions by integrated Hessians \cite{janizek2020explaining} and general neural interactions by aggregated weights \cite{tsang2017detecting}. However, all these methods are based on approximations and they cannot guarantee consistent interpretation results.  
	
	An important fact about ReLU DNNs is that the function expressed by a ReLU network is piecewise linear \cite{montufar2014number}.  One popular measure of expressivity is the total number of distinct activation patterns \cite{raghu2017expressive, serra2018bounding, hanin2019deep}, since an activation pattern is known to be associated with a convex polytope of local linear features. The use of activation pattern also appears in \cite{chu2018exact} (under the name of hidden neuron configuration) in study of piecewise linear neural networks with binary responses, for which a so-called exact interpretability is proposed to interpret the polytope boundary features. Besides, the activation pattern is  used recently by \cite{gopinath2019property} as a central tool to explain the neural network behavior in terms of input and layer properties.

	For deep ReLU networks that are usually over-parametrized, pruning is an effective way to simplify the network; see e.g.~\cite{frankle2018lottery} with ``lottery ticket hypothesis'' and  \cite{serra2020lossless} based on mixed-integer linear programming. For a large network, pruning is to fix certain entries of weight matrices to be zero, resulting in subnetworks with fewer unknown parameters. It is evident from multiple recent works that such pruned subnetworks may not only predict well without sacrifice on prediction performance but also enjoy the improved interpretability due to network simplification.  The interpretability or property verification of deep ReLU networks based on network simplification can be also referred to as \cite{gopinath2019property, katz2017reluplex}.


	\subsection{Main Contributions}
	Our purpose of this study is to unwrap the black box of ReLU DNNs and endue them with intrinsic interpretability. We develop a convenient toolkit called \aletheia for interpretability, diagnostics, and simplification of deep ReLU networks. Our main contributions in this paper are summarized as follows. 
	\begin{enumerate}
		\item An unwrapper is developed to disentangle a pre-trained ReLU DNN into an equivalent set of local linear models (LLMs). Such unwrapper for both regression and classification problems is rigorously supported by local linear representation of ReLU DNNs based on the device of activation pattern and activation region.  
		\item LLM-based intrinsic interpretability is developed for deep ReLU networks, including local linear profile and joint importance of an individual feature, simultaneous interpretation of multiple LLMs through the parallel coordinate plot, and local interpretability by statistical inference. 
		\item LLM-based diagnostics are developed for deep ReLU networks for checking the sizes of local regions, the extrapolation strengths of LLMs, and the similarities among the LLM coefficients. Single-instance or single-class regions are found to be a critical issue for over parametrized ReLU DNNs.   
		\item An effective merging algorithm is proposed for LLM-based network simplification. It combines the locally homogeneous LLMs with similar coefficients subject to local region connectivity. The single-instance or single-class regions are merged to nearest neighbor LLMs with large/medium sample sizes.  Besides, an optional flattening strategy is proposed to further simplify the network. 
		\item A new Python package called  \aletheia is developed as a convenient toolkit for unwrapping the black box of deep ReLU networks. It includes efficient algorithms for local linear representation and network simplification, as well as visualization tools for interpretation and diagnostics. The use of \aletheia is demonstrated through simulation examples, benchmark datasets, and a real case study in credit risk assessment.
	\end{enumerate}
	

	The rest of the paper is organized as follows. In Section~2 we derive the local linear representation for deep ReLU networks based on the activation pattern and activation region. Then Section~3 presents the LLM-based interpretability, and Section~4 discusses LLM-based diagnostics. In Section~5 we develop the merging and flattening techniques for network simplification. Section~6 presents a real case study in home lending credit risk assessment, as an illustration of the proposed methodology in real-life environments.  Section~7 wraps up the paper with concluding remarks.

	\section{Local Linear Representation}\label{sec:LLR}
\begin{figure}[htp!]
\cn{\includegraphics[width=0.75\textwidth]{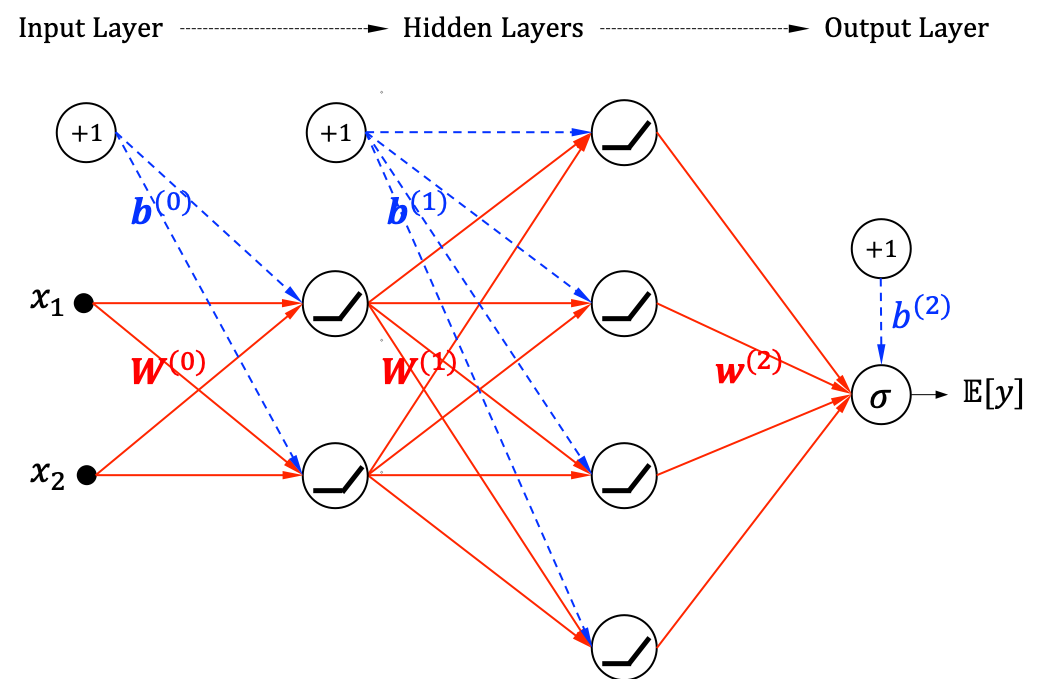}} 
\caption{Toy ReLU Net with two hidden layers of node sizes $[2,4]$. The network parameters $\thth$ include the weights $\W^{(0)}_{2\times 2}, \W^{(1)}_{4\times 2},  \w^{(2)}_{1\times 4}$ and the biases 
$\b^{(0)}_{2\times 1},  \b^{(1)}_{4\times 1}, b^{(2)}_{1\times 1}$.}\label{fig:DemoReLU_Net}
\end{figure}

	Denote by $\NN$ a feedforward ReLU network that has the input layer with $\x \in \Omega \subseteq\RRR^d$ and $L$ hidden layers with neuron  sizes $[n_1, \ldots, n_L]$. For each hidden neuron $\u_i^{(l)}$, denote its input (from left) as $z_i^{(l)}$ and the output (to right) as $\chi_i^{(l)}$, then by the ReLU activation
	\begin{equation}\label{ReLU}
	\chi_i^{(l)} = \max\{0, z_i^{(l)} \}, 
	\ \ \mbox{ for } i=1,\ldots,n_l \mbox{ and }  l=1,\ldots,L.
	\end{equation}
	From layer $l-1$ to $l$, given the weight matrix $\W^{(l-1)}$ of size $n_{l}\times n_{l-1}$ and the bias vector $\b^{(l-1)}$ of length $n_{l}$,  the layer-$(l-1)$ output $\bm\chi^{(l-1)}$ leads to the layer-$l$ input $\z^{(l)}$ by the affine transformation 
	\begin{equation}\label{affine}
	\z^{(l)} = \W^{(l-1)} \bm\chi^{(l-1)} + \b^{(l-1)}, \ \ \mbox{for }  l=1, \ldots, L,
	\end{equation}
where the layer zero corresponds to the input layer with $\bm\chi^{(0)} = \x$, $\W^{(0)}$ of size $n_1\times d$, and $\b^{(0)}$ of length $n_1$. As for the output layer (i.e. layer $L+1$), the layer-$L$ output $\chi_i^{(L)}$ can be used to predict the final response $y$ by the generalized linear model (GLM), 
	\begin{equation}\label{GLM}
	\EEE[y]  = \sigma(\w^{(L)} \bm\chi^{(L)} + b^{(L)}) \equiv \sigma(\eta(\x)) 
	\end{equation}
	where the $\sigma$ function can be the identity link for a regression problem (i.e. $y$ is continuous) or the logit link for a classification problem (i.e. $y$ is binary). In such GLM formulation with univariate response,  $ \W^{(L)}  \equiv \w^{(L)}$ is a row vector and $ \b^{(L)} \equiv b^{(L)}$ is a scalar.  
For convenience, let us use $\thth$ to denote the set of network parameters $\{\W^{(l)}, \b^{(l)}, ~ l=0,1,\ldots, L\}$. 
	See Figure~\ref{fig:DemoReLU_Net} for a toy ReLU net with bivariate inputs, two hidden layers with node sizes $[2,4]$, and a univariate output. Note that when $y$ is one-hot encoded (either binary or  multi-category classification), we can use the softmax link with a straightforward extension. 

\subsection{Activation Pattern}
For each hidden neuron, the ReLU activation (\ref{ReLU})  has two states, namely ``on'' (active, when $z\geq 0$) and ``off'' (inactive, when $z<0$). To investigate the on/off states of all hidden neurons in the ReLU network, we employ the notion of activation pattern \cite{raghu2017expressive} and formally define it as follows. 
	
\begin{definition}[Activation Pattern] For $\NN$ with $L$ hidden layer sizes $[n_1, \ldots, n_L]$, define the activation pattern to be the following binary vector 
\begin{equation}\label{Pattern}
\P=[\P^{(1)}; \ldots; \P^{(L)}] \in \{0, 1\}^{\sum_{i=1}^L n_l}
\end{equation}
which indicates the on/off state of each hidden neuron in the network. Specifically, the component $\P^{(l)}$ is called a layered pattern for $l=1,\ldots,L$. The activation pattern $\P$ is said to be  trivial if there is at least one $\P^{(l)} \equiv 0$ for some $l$. 
	\end{definition}
	
	The length of activation pattern is $\sum_{i=1}^L n_l$, which equals the total number of hidden neurons in the network.  For each input $\x\in\Omega$, through the feedforward neural network with fixed parameters, it would determine the values of $\{\z^{(l)}, \bm\chi^{(l)}\}$ by (\ref{affine}) and (\ref{ReLU}) for $l=1,\ldots,L$ sequentially. Each $\z^{(l)}$ would determine the on/off states of the ReLU neurons at layer $l$, therefore determine a layered pattern, as denoted by $\P^{(l)}(\x)$. Thus, any input instance $\x$ corresponds to a particular activation pattern of the form
	\begin{equation}\label{Pattern_x}
	\P(\x) = [\P^{(1)}(\x); \ldots; \P^{(L)}(\x)],
	\end{equation}
which can be either trivial or non-trivial. We are now ready to define the {\em activation region} in association with each distinct pattern 
\begin{equation}\label{ActRegion}
	\RR^{\P} = \{\x\in\Omega: \P(\x) = \P\},
\end{equation}
which is known to be a convex polytope with closed-form boundaries to be discussed below.

\begin{figure}[htp!]
\cn{\includegraphics[width=0.36\textwidth]{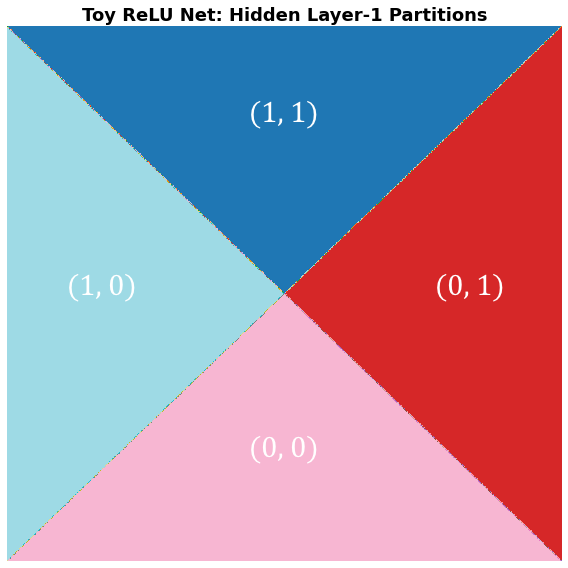}
\quad\quad \includegraphics[width=0.36\textwidth]{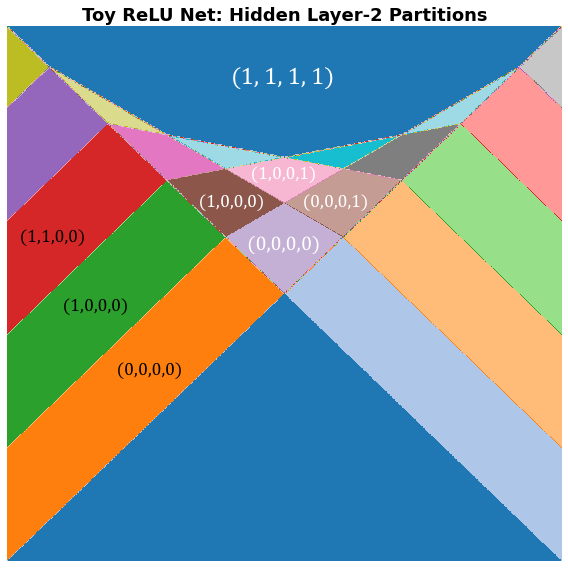}} 
\caption{Region partitions by the toy ReLU net with tailor-made parameters (\ref{DemoPara}).}\label{fig:DemoActRegion}
\end{figure}

The activation region plays an important role in understanding how the hidden layers/nodes partition the full space $\Omega$ into disjoint sub-spaces. To illustrate how it works, consider the toy ReLU net in Figure~\ref{fig:DemoReLU_Net} with the following tailor-made weight and bias parameters for the two hidden layers,
\begin{equation}\label{DemoPara}
\W^{(0)} = \frac{1}{\sqrt{2}}
\begin{pmatrix}
-1 & 1 \\
1 & 1 
\end{pmatrix}
, \ 
\b^{(0)} = 
\begin{pmatrix}
0  \\
0  
\end{pmatrix}
; \quad  
\W^{(1)} = 
\begin{pmatrix}
1 & 1/4 \\
1/2 & 1/3\\
1/3 & 1/2\\
1/4 & 1 
\end{pmatrix}
, \ 
\b^{(1)} = \frac{3}{10}
\begin{pmatrix}
-1  \\
-1 \\
-1\\
-1 
\end{pmatrix}.
\end{equation}
In this example, the affine transformation  $\z^{(1)} = \W^{(0)}\x + \b^{(0)}$ is an orthogonal rotation. It is easy to check the boundary conditions between on/off states of two hidden nodes in the first layer, given by
$$
\left\{\begin{aligned}
\mbox{on: } & -x_1 + x_2 > 0 \\
\mbox{off: } & -x_1 + x_2 \leq 0
\end{aligned}\right.
 \ \mbox{ and } \  
 \left\{\begin{aligned}
\mbox{on: } & x_1 + x_2  > 0 \\
\mbox{off: } & x_1 + x_2 \leq 0
\end{aligned}\right..
$$
See Figure~\ref{fig:DemoActRegion} (left) for the plot of such boundaries within the 2D domain $[-1, 1]^2$. The domain is divided into four sub-domains, each corresponding to a unique activation pattern (as annotated). More specifically, it includes three non-trivial patterns $(1, 1), (1, 0), (0, 1)$ and one trivial pattern $(0,0)$.

Now consider the forward propagation from the first hidden layer to the second hidden layer. Among the four regions shown in Figure~\ref{fig:DemoActRegion} (left), the bottom trivial $(0,0)$-quadrant receives constant inputs $\z^{(2)}$, by $\bm\chi^{(1)} = \zero$ and  equation (\ref{affine}); so this region generates the same constant output thereafter.  On the other hand, for the top $(1,1)$-quadrant, the input is updated to be $\bm\chi^{(1)} = \z^{(1)}$. Within this region,  the internal boundary conditions between on/off states of four hidden nodes can be determined by the affine-transformed variables $\z^{(2)} = \W^{(1)}\z^{(1)} + \b^{(1)}$, i.e.
$$
\begin{aligned}
\mbox{Node 1: ~~} & ~~z^{(1)}_1 + \frac{1}{4}z^{(1)}_2 -\frac{3}{10}  & > 0 \mbox{ or } \leq  0  \\
\mbox{Node 2: ~~} & \frac{1}{2} z^{(1)}_1 + \frac{1}{3}z^{(1)}_2 -\frac{3}{10} & > 0 \mbox{ or } \leq  0  \\
\mbox{Node 3: ~~} & \frac{1}{3} z^{(1)}_1 + \frac{1}{2}z^{(1)}_2 -\frac{3}{10} & > 0 \mbox{ or } \leq  0  \\
\mbox{Node 4: ~~} & \frac{1}{4} z^{(1)}_1 + ~~z^{(1)}_2 -\frac{3}{10}  & > 0 \mbox{ or } \leq  0  
\end{aligned}
$$
These four inequalities divide the domain into 11 sub-regions in total, as visualized in the top quadrant of Figure~\ref{fig:DemoActRegion} (right)\footnote{Note that this visualization is similar to Figure~2 in \cite{serra2018bounding} who used a ReLU net with 3 hidden layers of [2,2,2] nodes. However, there is subtile difference in the piecewise linear cuts. Both using 6 hidden nodes, \cite{serra2018bounding} obtained 20 regions in total (which is less than the full potential expressivity for a depth-4 network), and we obtained 22 regions in total (which achieves the full potential expressivity for a depth-3 network).}. For each of the 11 sub-regions, its corresponding activation pattern can be derived from the on/off states of hidden nodes. Due to small space sizes, only five sub-regions are annotated (in white-colored text) with explicit activation patterns. Similarly, we can analyze the left $(1,0)$-quadrant and the right $(0,1)$-quadrant and obtain their partitions by the four hidden nodes.  In the above inequalities, muting $z^{(1)}_2$ for the $(1,0)$-region (or muting $z^{(1)}_1$ for the $(0,1)$-region) leads to the internal boundary conditions, as well as the corresponding activation patterns. In Figure~\ref{fig:DemoActRegion} (right), the left $(1,0)$-quadrant is divided into 5 sub-regions, for which we annotate (in black-colored text) three examples of layered activation patterns.  

By this toy example, it is clear that the original input space is sequentially partitioned into multiple activation regions, and each region is associated with a unique activation pattern of the hidden nodes in each layer. It is also obvious that each activation region is a convex polytope, since all the boundary conditions are linear inequalities. In the meanwhile, it is important to observe from this example that the toy ReLU net generates the activation patterns with certain rules; in another word, an arbitrary binary pattern is not always expressible. For an instance, $(1, 0; 0, 0, 1, 1)$ is not a legitimate activation pattern in Figure~\ref{fig:DemoActRegion}. 
 
For ReLU DNNs in general, we have the following lemma that has partially appeared before in \cite{raghu2017expressive, hanin2019deep}. We use $\PP_{\rm expr}(\NN)$ to denote the set of distinct activation patterns that are expressible by a ReLU net $\NN$.

\begin{lemma}[Convex Partitioning]\label{lem:ActRegion}
For a ReLU DNN $\NN$,  each activation pattern $\P=[\P^{(1)}; \ldots; \P^{(L)}]$ in $\PP_{\rm expr}(\NN)$ is associated with a convex activation region $\RR^{\P}  \subset\Omega$, satisfying the following inequality constraints based on the layer-wise  affine-transformed variables:
\begin{equation}
(-1)^{\P^{(l)}} \circ \z^{(l)} \leq \zero, \quad l = 1, \ldots, L
\end{equation}
where $\circ$ denotes the Hadamard product (element-wise).  Moreover, for any two distinct activation patterns, their corresponding activation regions are disjoint. Hence, the ReLU net $\NN$ partitions the original input space into a finite set of convex sub-spaces, i.e., 
\begin{equation}
\Omega = \bigcupdot_{\P\in\PP_{\rm expr}(\NN)} \RR^{\P}.
\end{equation}
\end{lemma}

In Lemma~\ref{lem:ActRegion}, $\PP_{\rm expr}(\NN)$ is yet to be determined. This has been an active topic in the expressivity study of DNNs, in particular the upper and lower bounds of the total number of distinct patterns \cite{montufar2014number, raghu2017expressive, serra2018bounding, hanin2019deep}. According to \cite{raghu2017expressive}, the cardinality of $\PP_{\rm expr}(\NN)$ can be as large as $O(k^{dL})$ for a ReLU net with $L$ hidden layers of width $k$.  In this paper, our interest is to interpret a pre-trained network $\NN_{\rm train}$ based on training dataset $\XX_{\rm train}$, for which the useful activation patterns (to be unwrapped in Section~\ref{sec:unwrap}) turns out to be much fewer.

\subsection{Local Linear Models} 
A ReLU DNN partitions the input space into many sub-regions, and each sub-region is determined by a unique activation pattern. By multi-layer propagation, the original input features are sequentially transformed by (\ref{affine}) subject to the ReLU activation (\ref{ReLU}). Finally, the prediction takes the linear form
\begin{equation}\label{FinalLP}
\eta(\x) = w^{(L)} \bm\chi^{(L)} + b^{(L)}, 
\end{equation}
up to a pre-specified $\sigma$ link function. In this section, we derive the final features $\bm\chi^{(L)}$ explicitly for each sub-region and obtain the local linear models in closed form.

	For a trivial activation pattern when at least one layer has all inactive neurons, the output of such a layer would become all zero. For the corresponding trivial regions (e.g. the sub-regions annotated with $(0,0)$ and $(0,0,0,0)$'s in Figure~\ref{fig:DemoActRegion}), the prediction would be a constant value within each region, since the final output is only affected by the layer-wise bias terms, but not the original input variables $\x$. 	

	For a non-trivial activation pattern $\P=[\P^{(1)}; \ldots; \P^{(L)}]$, let us introduce a diagonal matrix $\D^{(l)}$ for each layer, whose diagonal takes the same $(0,1)$-entries as $\P^{(l)}$, i.e., 
	$$
	\D^{(l)} = \mbox{diag}(\P^{(l)}), \quad\mbox{for } l=1,\ldots,L.
	$$  
Using this notation, the layer-wise output after ReLU activation (\ref{ReLU}) can be effectively expressed in vector form as the following chain rule,
\begin{equation}\label{DiagChain}
\bm\chi^{(l)} =  \max\{\zero, \z^{(l)} \} = \D^{(l)}\z^{(l)} = \D^{(l)}\big(\W^{(l-1)}\bm\chi^{(l-1)} + \b^{(l-1)}\big).
\end{equation}
Thus, we may derive the final features $\bm\chi^{(L)}$ recursively as follows,
	\begin{eqnarray*}
		\bm\chi^{(L)} & = &   \D^{(L)}\big(\W^{(L-1)}\bm\chi^{(L-1)} + \b^{(L-1)}\big) \\
		& = &   \D^{(L)}\W^{(L-1)}  \D^{(L-1)}\big(\W^{(L-2)}\bm\chi^{(L-2)} + \b^{(L-2)}\big)  + \D^{(L)}\b^{(L-1)}\\
		& = & \cdots \\
		& = &  \D^{(L)}\W^{(L-1)}\cdots \D^{(1)}\W^{(0)}  \x + \sum_{l=1}^{L-1} \D^{(L)}\W^{(L-1)}\cdots \D^{(l+1)}\W^{(l)} \D^{(l)}\b^{(l-1)} + \D^{(L)}\b^{(L-1)}\\
		& = & \prod_{h=1}^L \D^{(L+1-h)} \W^{(L-h)} \x + \sum_{l=1}^{L-1} \prod_{h=1}^{L-l} \D^{(L+1-h)} \W^{(L-h)}\D^{(l)}\b^{(l-1)} + \D^{(L)}\b^{(L-1)}.
	\end{eqnarray*}
Furthermore, the linear prediction (\ref{FinalLP}) by the output layer is given explicitly by 
$$
\eta(\x) = \prod_{h=1}^L \W^{(L+1-h)}  \D^{(L+1-h)} \W^{(0)} \x  + 
\sum_{l=1}^{L}\prod_{h=1}^{L+1-l} \W^{(L+1-h)}  \D^{(L+1-h)} \b^{(l-1)} + b^{(L)},  
$$
in which $\W^{(L)} \equiv \w^{(L)}$ for the notation convenience. Hence, we have the following 
theorem that characterizes deep ReLU networks via local linear models (LLMs). 
	
\begin{theorem}[Local Linear Model]\label{thm:LLM}
For a ReLU DNN and any of its expressible activation pattern $\P$,  the local linear model  on the activation region  $\RR^{\P}$ is given by
\begin{equation}\label{LLM}
\eta^{\P}(\x)  = \wtilde^{\P}\x  + \tilde{b}^{\P}, \quad \forall \x\in \RR^{\P}
\end{equation}
with the following closed-form parameters 
\begin{equation}\label{LLM_coef}
\wtilde^{\P} = \prod_{h=1}^L \W^{(L+1-h)}  \D^{(L+1-h)} \W^{(0)}, \quad
\tilde{b}^{\P} = \sum_{l=1}^{L}\prod_{h=1}^{L+1-l} \W^{(L+1-h)}  \D^{(L+1-h)} \b^{(l-1)} + b^{(L)}. 
\end{equation}
\end{theorem}

By combining the region partitioning in Lemma~\ref{lem:ActRegion} and the local linear models in Theorem~\ref{thm:LLM}, we have a local linear representation for ReLU DNNs. In another word, a deep ReLU network can be equivalently re-expressed as a finite set of local linear models each functioning exclusively on a disjoint convex sub-region. It is clear that the activation pattern plays a critical role in such local linear representation. A remaining question is how to determine the activation patterns for network interpretation purpose, for which we develop an effective unwrapper for pre-trained ReLU DNNs.

\subsection{Unwrapping Pre-trained DNNs}\label{sec:unwrap}
\begin{figure}[htp!]
\cn{\includegraphics[width=0.45\textwidth]{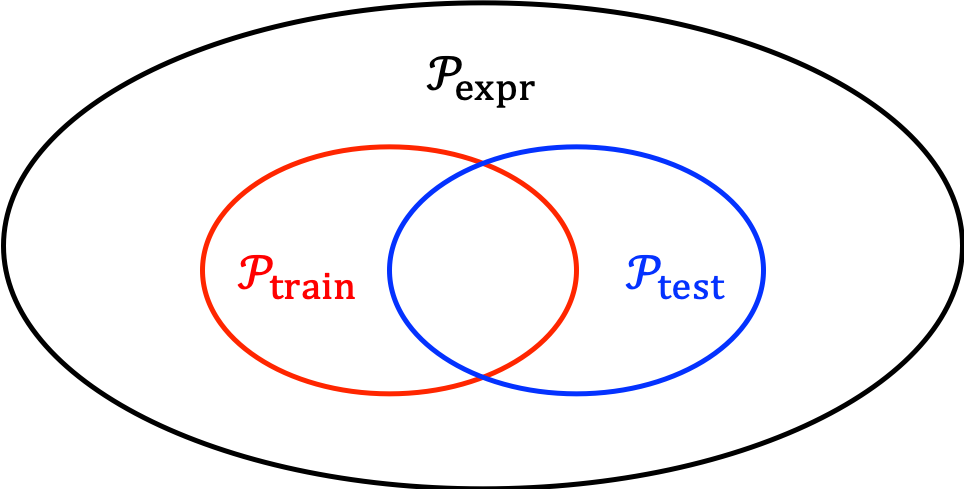}} 
\caption{Sets of activation patterns: theoretically expressible $\PP_{\rm expr}$, training data induced $\PP_{\rm train}$ and testing data induced $\PP_{\rm test}$.}\label{fig:SetPattern}
\end{figure}

For a generic ReLU net $\NN$, the number of expressible activation patterns as in $\PP_{\rm expr}(\NN)$ grows exponentially in network depth, width and input dimensionality. It is just unrealistic to investigate each possible activation pattern, and unnecessary, too. In this paper, we concentrate on the pre-trained DNNs $\NN_{\rm train}$ based on training data $\XX_{\rm train}$, for which we define the set of training data induced activation patterns:
\begin{equation}\label{SetPattern_train}
\PP_{\rm train} = \{\P\in\PP_{\rm expr}: \RR^{\P} \cap \XX_{\rm train}  \neq \emptyset \},
\end{equation}
and similarly we can define the set of testing data induced activation patterns $\PP_{\rm test} = \{\P\in\PP_{\rm expr}: \RR^{\P} \cap \XX_{\rm test}  \neq \emptyset \}$. It simply means there is at least one training (resp. testing) instance that will be covered by the $\P$-associated activation region for $\P\in \PP_{\rm train}$ (resp. $\P\in \PP_{\rm test}$).  Their relationship is depicted in Figure~\ref{fig:SetPattern}. 
Meanwhile, denote the {\em activation set} of training instances for each activation pattern by
\begin{equation}\label{Region_train}
	\RR_{\rm train}^{\P} = \{\x\in\XX_{\rm train}: \P(\x) = \P\}, \  \mbox{ for } \P\in\PP_{\rm train}
\end{equation}
and similarly the activation set of testing instances by $\RR_{\rm test}^{\P}$. Obviously,  $\RR_{\rm train}^{\P} = \emptyset$ if $\P\notin \PP_{\rm train}$. Note that although $\RR_{\rm train}^{\P}$  can be viewed as as subset of the convex region $\RR^{\P}$, the set $\RR_{\rm train}^{\P}$ is a discrete set of instances without formal convex property. 

By Theorem~\ref{thm:LLM}, the pre-trained ReLU net $\NN_{\rm train}$ for the training instances can be equivalently represented by the following set of LLMs, 
	\begin{equation}\label{SetLLM} 
	\LL_{\rm train} =  \left\{\eta^{\P}(\x)  = \wtilde^{\P} \x + \tilde{b}^{\P}, \ \ \mbox{for } \x \in \RR_{\rm train}^{\P},  \  \P\in \PP_{\rm train}\right\}
	\end{equation}
where the LLM coefficients $\{\wtilde^{\P}, \tilde{b}^{\P}\}$ are given by (\ref{LLM_coef}). Such a formal procedure for determining $\big\{\LL_{\rm train}, \PP_{\rm train}, \{\RR_{\rm train}^{\P}: \P\in\PP_{\rm train}\}\big\}$ is called an {\bf unwrapper} for pre-trained ReLU DNNs. 
Algorithmically, the unwrapper may identify distinct activation patterns in $\PP_{\rm train}$ by individual training instances\footnote{For large-scale training data (millions or more sample sizes), certain smart sampling strategy can be employed for computational speed up. This is currently under our investigation.}, for which all $\sum_{l=1}^L n_l$ hidden nodes need to be checked by layer-wise propagated $\z^{(l)}$ for each $\x\in\XX_{\rm train}$. In this way, each identified activation pattern $\P$ comes with its supporting instances, i.e. $\RR_{\rm train}^{\P}$.  Finally, each LLM in $\LL_{\rm train}$ can be evaluated for each unique $\P\in \PP_{\rm train}$ straightforwardly.
	
A new Python package called \aletheia is under our development, which includes {\sf UnwrapperClassifier} and {\sf UnwrapperRegressor} of ReLU nets for both classification and regression problems. This unwrapper provides a solid foundation for local linear interpretability, diagnostics, and simplification of deep ReLU networks in the next sections.

\section{LLM-based Network Interpretability}\label{sec:interpret}
The local linear representation by Theorem~\ref{thm:LLM} enables us to directly interpret ReLU DNNs based on LLMs. In this section, we first introduce the local linear profile for the purpose of interpreting individual feature importances and partial relationships with final prediction, then discuss how to interpret the joint effects of LLMs through effective use of parallel coordinate plot. 

	For better illustration of the proposed methods, we use the following one-, two- and multi-dimensional examples under both regression and classification settings:
	\begin{itemize}
		\item {\bf ChirpWave} (regression): $y_i=\sin(2\pi/(x_i+0.2))+\ve$ for $x_i\in[0,1], ~ i=1,\ldots, 2000$, where the i.i.d. noise is simulated by $\ve\sim N(0,0.1^2)$;
		\item {\bf CoCircles} (binary classification):  concentric circle dataset with two features, which can be generated by {\sf sklearn.datasets.make\_circles(n\_samples=2000, noise=0.1)}; 
		\item {\bf BostonHouse} (regression):  Boston house prices dataset with 506 samples and 13 features, which can be loaded by  {\sf sklearn.datasets.load\_load\_boston()}; 
		\item {\bf FicoHeloc} (binary classification):  9871 samples with 35 features after preprocessing the raw data from the FICO website\footnote{FICO 2018 xML challenge: https://community.fico.com/s/explainable-machine-learning-challenge}. 
	\end{itemize}
	Each dataset is split into training (80\%) and testing (20\%) sets upon random permutation. For simplicity, we use {\sf sklearn.neural\_network} (either {\sf MLPRegressor} or {\sf MLPClassifier}) for implementing the ReLU net with four hidden layers each with 40 neurons, except for the small BostonHouse data with two hidden layers each with 20 neurons. Each ReLU net is trained for 2000 epochs using the Adam optimizer, together with an early stopping rule when the validation performance (20\% of training data) does not get improved for 100 epochs. 
	
	\begin{table}[hbtp!]
		\centering
		\renewcommand\tabcolsep{8pt}
		\renewcommand\arraystretch{1.2}
		\caption{Four datasets and deep ReLU network model performances. Each dataset is split into training (80\%) and testing (20\%) sets. The ReLU net size stands for the hidden layer architecture. The number of LLMs stands for the cardinality of unwrapped $\CC_{\rm train}$  from the fitted $\NN_{\rm train}$ based on the training data.  The training and testing performances stand for MSE (for ChirpWave and BostonHouse) and AUC (CoCircles and FicoHeloc). } \label{tab:DnnPerf}
		\begin{tabular}{l|rrrrr}
			\hline\hline  \vspace{-0.1cm}
			Dataset 		& Data Size 	& ReLU~Net Size  &  Num.~LLMs  &   Training & Testing \\\hline
			ChirpWave	& 2000 x 1 	&   [40]*4 		&  64		&  0.0113	& 0.0117  \\
			CoCircles   	& 2000 x 2 	&   [40]*4 		&  530  	&  0.9215	& 0.9320    \\ 
			BostonHouse  	& 506 x 13	&   [20]*2 		&  155 	&  0.0022 	& 0.0094   \\
			FicoHeloc 	& 9871 x 35	&   [40]*4 		&  7423  	&  0.8325 & 0.7980\\
			\hline\hline
		\end{tabular}
	\end{table}

	By the unwrapper introduced in the previous section,  each trained ReLU net $\NN_{\rm train}$ can be re-expressed as a set of LLMs $\LL_{\rm train}$ together with the corresponding set of activation patterns $\PP_{\rm train}$. Note that the total number of LLMs (i.e. cardinality of $\PP_{\rm train}$) depends not only on the training data but also on the local optimal solution trained by SGD optimization. In the meanwhile, we obtain the training and testing performances (MSE for regression problems and AUC for classification problems). These details are listed in Table~\ref{tab:DnnPerf}.

\subsection{Local Linear Profile and Joint Importance}
	To study the marginal effect of each dimensional feature, we define the local linear profile and the joint importance measure based on the unwrapped LLMs.  
	
	\begin{definition}[Local Linear Profile]\label{defn:LLProfile}
		For a pre-trained $\NN_{\rm train}$ re-expressed by the set of LLMs (\ref{SetLLM}), we define the local linear profile of each dimensional feature $x_j~(j=1,\ldots,d)$ by the set of local marginal linear functions upon centering
		\begin{equation}\label{LLProfile}  
		\left\{\eta^{\P}(x)  = \tilde{w}^{\P}_j x - \frac{1}{|\RR^{\P}_{\rm train}|}\sum\nolimits_{\x_i\in \RR^{\P}_{\rm train}}\tilde{w}^{\P}_j x_{ij} , \  \P\in \PP_{\rm train}\right\},
		\end{equation}
		and define the local intercept profile by the set of local bias terms  $\big\{\tilde{b}^{\P}, \  \P\in \PP_{\rm train}\big\}$. Furthermore, we define the joint importance of intercept and individual features by 
		\begin{equation}\label{JI}
		\begin{gathered}
		{\rm JI}(b_0) =  \frac{\sum_{\P\in \PP_{\rm train}}|\RR^{\P}_{\rm train}| \big(\tilde{b}^{\P}\big)^2}{T}  \ \  \mbox{($b_0$: intercept)}\\
		{\rm JI}(x_j) =  \frac{\sum_{\P\in \PP_{\rm train}} |\RR^{\P}_{\rm train}| \big(\tilde{w}^{\C}_j\big)^2}{T}, \ \  j = 1, \ldots, d \ 
		\end{gathered}
		\end{equation}
		in which $T= \sum_{\P\in \PP_{\rm train}}|\RR^{\P}_{\rm train}| \big(\tilde{b}^{\P}\big)^2 + \sum_{j=1}^d\sum_{\P\in \PP_{\rm train}} |\RR^{\P}_{\rm train}|  \big(\tilde{w}^{\P}_j\big)^2$. 
	\end{definition}
	
\begin{figure}[htp!]
		\cn{\includegraphics[width=0.45\textwidth]{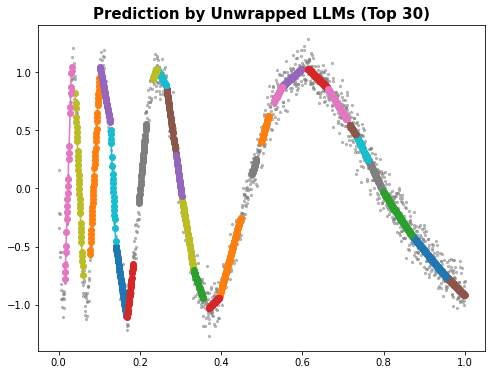}
			\ \ \	\includegraphics[width=0.43\textwidth]{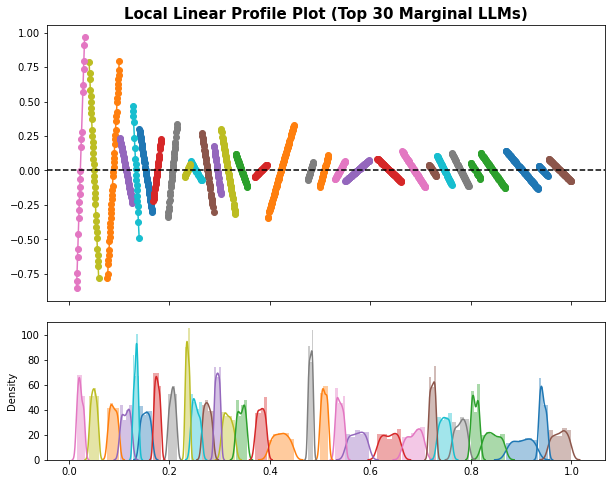}} 
		\caption{DNN prediction and local linear profile plot (Data: ChirpWave; ReLU Net: [40]*4)}\label{fig:Ex1Profile}
	\end{figure}

	By Defintion~\ref{defn:LLProfile}, it is straightforward to check the partial dependence of DNN prediction on each individual feature. We introduce a local linear profile plot, which visualizes the top marginal LLMs (ranked by the cardinality $|\RR_{\rm train}^{\P}|$) from (\ref{LLProfile}),  together with the sample distributions of the corresponding regions. Such a profile plot consists of multiple line segments each representing a marginal LLM upon centering. For the ChirpWave example, the DNN-predicted curve by the top-30 LLMs is shown in Figure~\ref{fig:Ex1Profile} (left), and the corresponding profile plot of their marginal LLMs is in Figure~\ref{fig:Ex1Profile} (right). Each line segment in the profile plot is supported by a small region without overlapping since all the marginal regions are disjoint in this one-dimensional case. Note that in the bottom part of the profile plot, the sample distribution for each activation region is separately smoothed by kernel density estimation upon normalization; so it does not reflect the total number of instances per region. 
	
\begin{figure}[htp!]
		\cn{\includegraphics[width=0.33\textwidth]{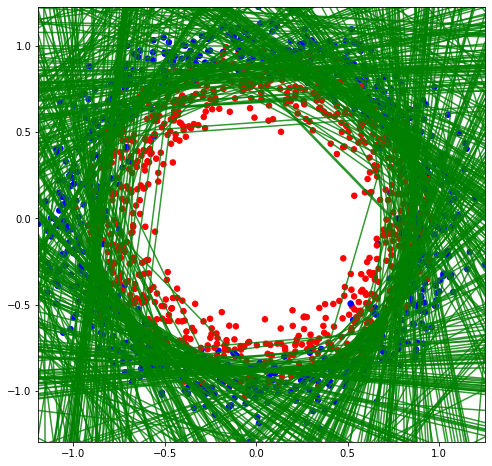}
			\ \ \	\includegraphics[width=0.31\textwidth]{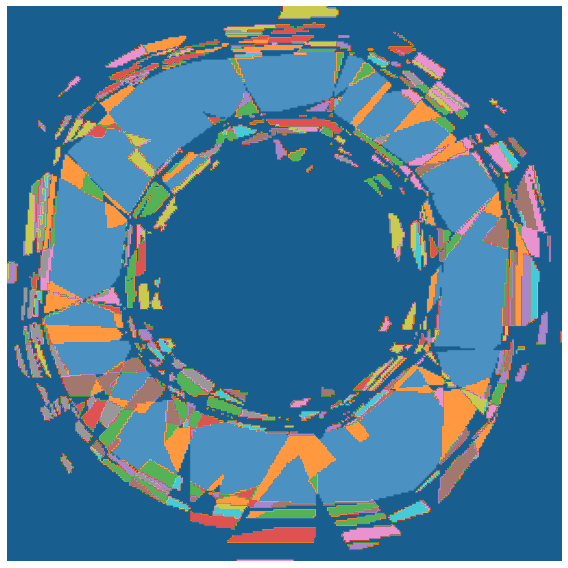}
			\includegraphics[width=0.32\textwidth]{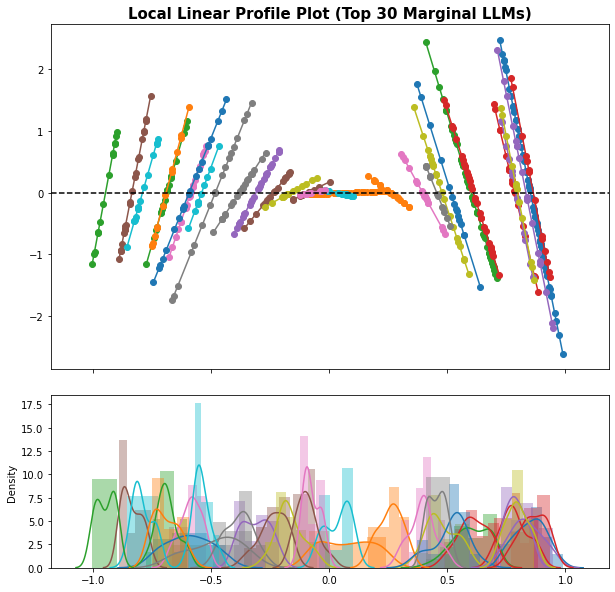}}
		\caption{Local linear models unwrapped from pre-trained ReLU net, the corresponding activation regions, and the local linear profile plot (Data: CoCircles; ReLU Net: [40]*4)}\label{fig: Ex2Profile}
	\end{figure}

	For the two-dimensional CoCircles example, the LLMs can also be visualized on a plane. See Figure~\ref{fig: Ex2Profile} (left) for the resulting 500+ LLMs unwrapped from the ReLU net. It is clear that these LLMs altogether approximate the underlying circle decision boundary in a piecewise linear manner. Each LLM corresponds to an activation region in the input space; as shown in Figure~\ref{fig: Ex2Profile} (middle). In this visualization, the large-connected blue regions represents the uncharted domain without training instances, for which no LLMs are unwrapped. 
The local linear profile plot (with top 30 marginal LLMs) for one of the features is shown in Figure~\ref{fig: Ex2Profile} (right).  In this case, the projections of local activation regions to a marginal feature may overlap with each other,  and the marginal LLMs may show the cross-over pattern. 
	

\begin{figure}[htp!]
		\cn{\includegraphics[width=0.45\textwidth, height=0.36\textwidth]{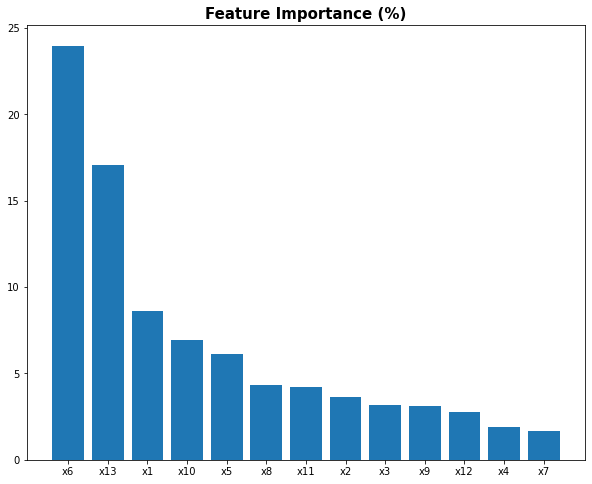}
			\ \ \ \includegraphics[width=0.45\textwidth]{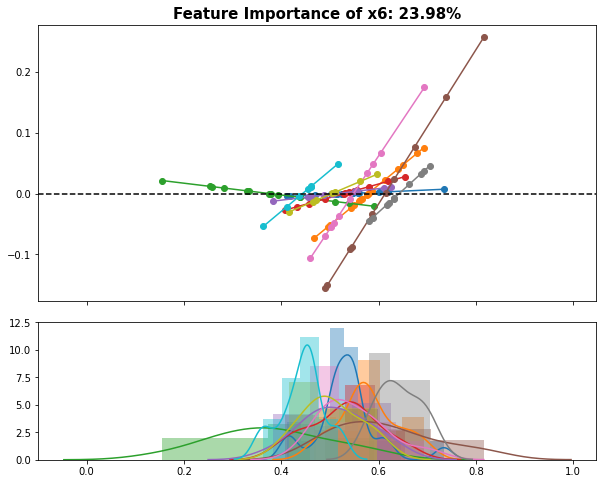}}
		\caption{Left: feature importance plot. Right: profile plot of the most important feature.  (Data: BostonHouse; ReLU Net: [20]*2)}\label{fig: Ex3Profile}

\bigskip	
		\cn{\includegraphics[width=0.45\textwidth, height=0.36\textwidth]{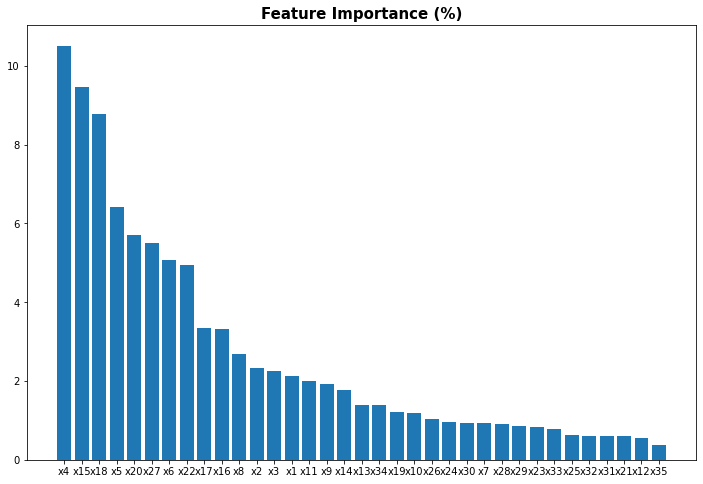}
			\ \ \ \includegraphics[width=0.45\textwidth]{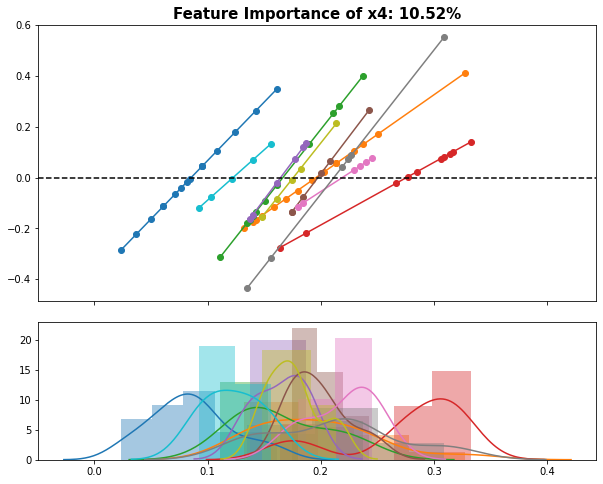}} 
		\caption{Left: feature importance plot. Right: profile plot of the most important feature.  (Data: FicoHeloc; ReLU Net: [40]*4)}\label{fig: Ex5Profile}
	\end{figure}

	For datasets with multi-dimensional features, according to Definition~\ref{defn:LLProfile}, we can quantify the joint importance of each feature and rank them in the decreasing order. See the bar charts of the sorted joint importance values for the BostonHouse and FicoHeloc examples in Figures~\ref{fig: Ex3Profile}--\ref{fig: Ex5Profile} (left).  After that, we can use the local linear profile plot to check the partial dependence for the identified important features. Figures~\ref{fig: Ex3Profile}--\ref{fig: Ex5Profile} (right) show the profile plots of the most important feature based on the top-10 LLMs in each example. 
	
	Thus, we have developed both the feature importance and profile plot for the main effect interpretation of pre-trained deep ReLU networks. This is similar to the post-hoc VI and PDP methods developed by \cite{breiman2001random, friedman2001greedy}.  However, as a key difference, our proposed method for ReLU DNNs is an intrinsic interpretability approach based on the unwrapped LLMs. 
	

\subsection{Parallel Coordinate Plot}
	It is also important to interpret the joint effects of multiple features simultaneously, which can be either difficult or easy. The joint interpretation is difficult if we want to rigorously quantify the two-factor or higher-order interaction effects in the functional ANOVA sense since the interaction effect varies locally from one region to another.  
	On the other hand, the joint interpretation is easy if we investigate the feature effects of each LLM separately, which actually involves only main effects within the activation region. In this paper, we take the easy approach while leaving the difficult one for our future work. 
	
\begin{figure}[htp!]
		\cn{\includegraphics[width=0.7\textwidth]{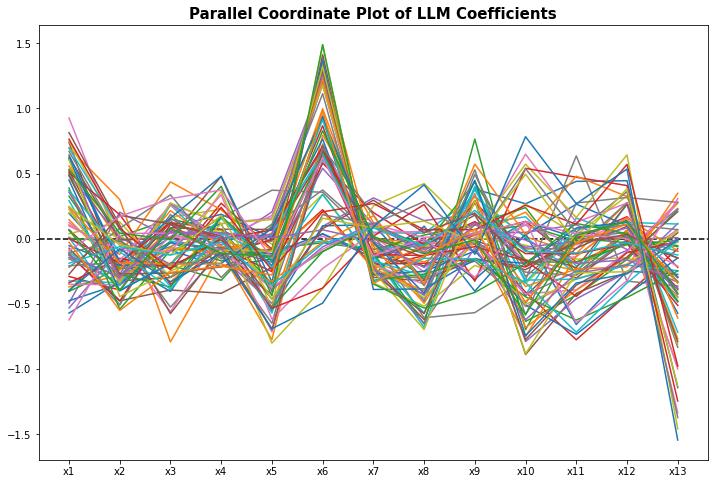}}
		\caption{Parallel coordinate plot of unwrapped LLMs excluding dummy cases (Data: BostonHouse; ReLU Net: [20]*2) }\label{fig:Ex3_PCplot}

\bigskip	
		\cn{\includegraphics[width=0.7\textwidth]{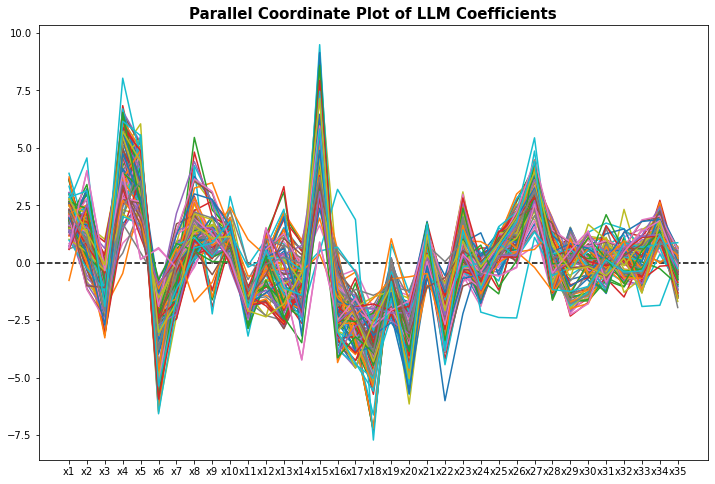}}
		\caption{Parallel coordinate plot of unwrapped LLMs excluding dummy cases  (Data: FicoHeloc; ReLU Net: [40]*4)}\label{fig:Ex5_PCplot}
	\end{figure}

	We suggest using the parallel coordinate plot for simultaneous visualization of LLM coefficients from different regions.  As a common way of visualizing high-dimensional data, the parallel coordinate plot treats each feature as a vertical axis and align all the features in natural ordering (by default) in parallel to each other. Here, each LLM is treated as a subject with its feature coefficients plotted as a connected line from left to right. Repeat it for multiple LLMs on the same chart,  and we obtain the graphical output; see Figures~\ref{fig:Ex3_PCplot}--\ref{fig:Ex5_PCplot} for the BostonHouse and FicoHeloc examples, where each colored line represents a different LLM with its feature coefficients\footnote{As an option, we may also include the intercept of each LLM as the starting vertical axis of the parallel coordinate plot.}.  Note that these plots are drawn after excluding the LLMs supported by single-instance or single-class regions, which can be identified if the region contains only the instances of the same response value. 
Later in Section~\ref{sec:diagnostics} when we discuss the network diagnostics, we will come back to such single-instance or single-class problems, which is a sign of overfitting with instability.  

	By the parallel coordinate plot, we may check the distribution of main effects from feature to feature and assess their effect consistency among multiple LLMs. For a particular feature, we may find that
	\begin{itemize}
		\item When most of its coefficients are large and have the same signs, it is implied that this feature is an important effect for the final prediction. Furthermore, this feature has a monotonic increasing effect if all coefficients are positive, and a monotonic decreasing effect if all coefficients are negative;  
		\item When most of its coefficients are large and have both positive and negative signs, it is implied that this feature has inconsistent slopes among local regions, which might be caused by either its own nonlinear effect or the interaction effects with other features;
		\item When most of its coefficients are small and close to zero, it is implied that this feature is not important and can therefore be excluded from the model. 
	\end{itemize}
	
	The joint interpretation by parallel coordinate plot is closely related to the feature importance and profile plot in the previous subsection. In an unrealistic special case when all the regions have the same number of training instances, the measure of joint importance (\ref{JI}) coincides with the sum of squared coefficients for each vertical axis in the parallel coordinate plot. Nevertheless, we can check the plots in Figures~\ref{fig:Ex3_PCplot}--\ref{fig:Ex5_PCplot} and find that a) {\sf x6, x13} are important features for the BostonHouse example, and b) {\sf x15, x4, x6, x18, x27} are important features for the FicoHeloc example. This is kind of consistent with the feature importance plot in Figures~\ref{fig: Ex3Profile}--\ref{fig: Ex5Profile} (left). Moreover, if the drill-down analysis for certain features is needed, we can use the local linear profile plot in Figures~\ref{fig: Ex3Profile}--\ref{fig: Ex5Profile} (right).

%
%
%
%
		
\section{LLM-based Network Diagnostics}\label{sec:diagnostics}
\begin{figure}[htp!]
		\cn{\includegraphics[width=0.8\textwidth]{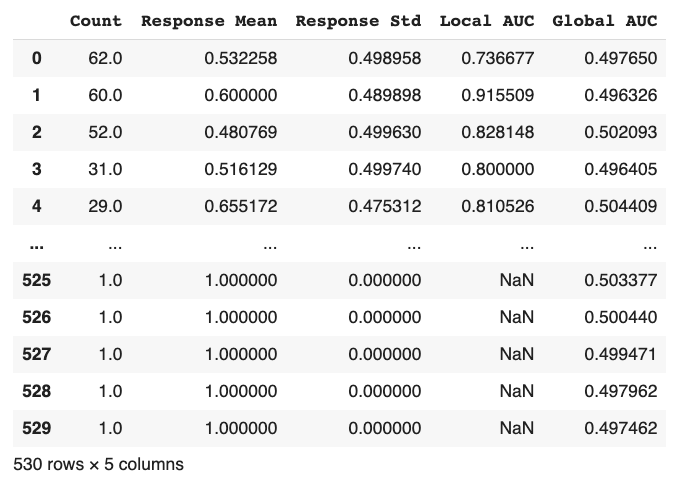}}
		\caption{Python code output from \aletheia: UnwrapperClassifier (Data: CoCircles)}\label{fig:Ex2_Output}
		
		\bigskip
		\cn{\includegraphics[width=0.8\textwidth]{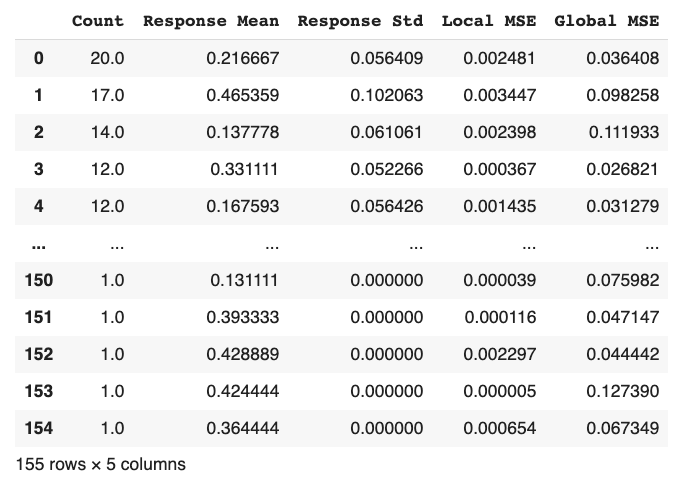}}
		\caption{Python code output from \aletheia: UnwrapperRegressor (Data: BostonHouse)}\label{fig:Ex3_Output}
	\end{figure}

	The unwrapper developed in Section~2 also enables us to perform diagnostics on a pre-trained ReLU DNN based on the set of LLMs. Here we try to address the following three questions: 
	\begin{itemize}
		\item[Q1.] Is it a large or small LLM (in terms of the support size)?
		\item[Q2.] Is it a strong or weak LLM (in terms of prediction performance)? 
		\item[Q3.] Is it a unique or duplicate LLM (in terms of model similarity)?
	\end{itemize}
	To answer these questions,   for each input of pre-trained ReLU net, the unwrapper in \aletheia is designed to generate numerical output that carries the following region-wise summary statistics,
	\begin{itemize}
		\item {\sf Count:} number of training instances within the activation region; 
		\item {\sf Response Mean:} average of response values in the region; 
		\item {\sf Response Std:} standard deviation of response values in the region; 
		\item {\sf Local MSE} (regression) or {\sf Local AUC} (classification):  performance on the training instances within the region; 
		\item {\sf Global MSE} (regression) or {\sf Global AUC} (classification):  performance on the entire training instances. 
	\end{itemize}
	The regions are sorted in the decreasing order of {\sf Count}.  See the unwrapper outputs for the CoCircles (classification) and BostonHouse (regression) examples in Figures~\ref{fig:Ex2_Output} and \ref{fig:Ex3_Output}. Our discussions about network diagnostics are based on such unwrapper outputs. 
	
	\subsection{Polar Coordinate Plot} 
	To answer Q1 and Q3, we propose a polar coordinate plot based on the region-wise count statistic and the corresponding LLM coefficients. For two-dimensional data, the bivariate coefficients upon unit norm scaling would determine the direction, while {\sf Count} determines the radius. For multi-dimensional LLM coefficients, we use the principal component analysis (PCA) for dimension reduction to PC1 and PC2, then draw the polar coordinate plot. 
	Besides, the blue color is used to mark the regions with {\sf Response Std = 0}, which corresponds to single-instance regions (regression) or single-class regions (classification). See the bottom part of Figures~\ref{fig:Ex2_Output} and \ref{fig:Ex3_Output} about these small regions. In the polar coordinate plot, these small blue-colored regions usually concentrate around the center of plot. 
	
\begin{figure}[htp!]
		\cn{\includegraphics[width=0.32\textwidth]{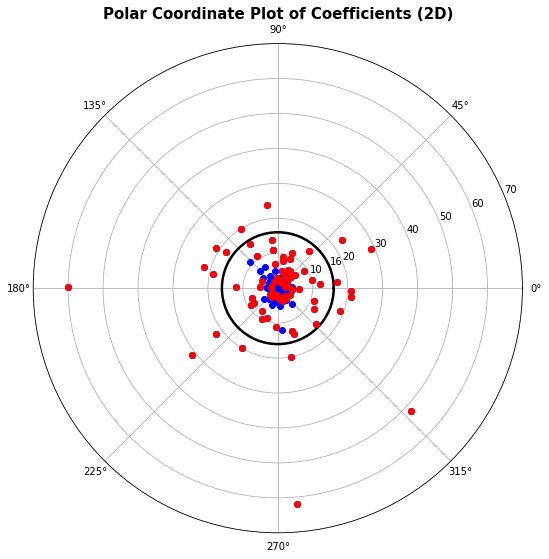}
			\ \ \includegraphics[width=0.32\textwidth]{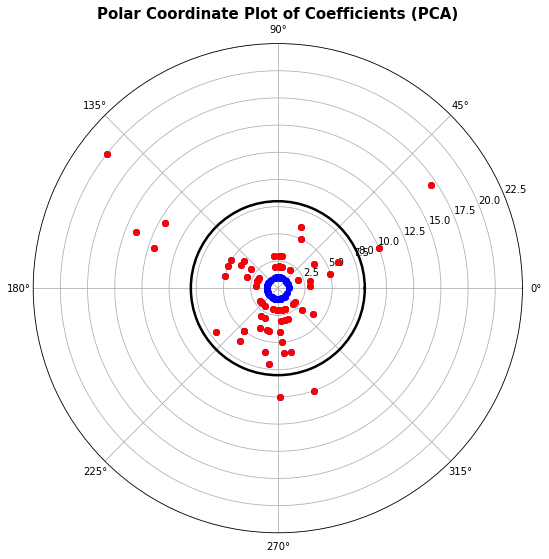}
			\ \ \includegraphics[width=0.32\textwidth]{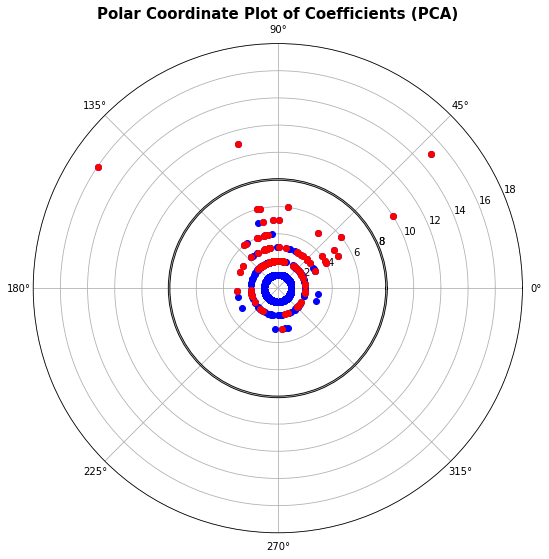}}
		\caption{Polar coordinate plot  with direction determined by LLM coefficients, while the radius determined by the count of instances.  The single-instance or single-class regions are marked in blue. 
			Left: CoCircles; Center: BostonHouse; Right: FicoHeloc.}\label{fig:Polar}
	\end{figure}

	See Figure~\ref{fig:Polar} for the polar coordinate plots for the CoCircles, BostonHouse, and FicoHeloc examples. It is found that the blue-colored regions occupy 84.9\%, 52.9\%, and 98.2\%, respectively. The proportions are very high, especially the two classification examples that involve a great amount of single-class regions.  As a matter of fact, the local interpretability discussed in Section~\ref{sec:interpret} is only applicable to large or medium regions exceeding the minimum sample size requirement, but not small regions with few instances.

	The polar coordinate plot is also helpful to check the similarity among large/medium LLMs. When multiple LLMs lie in approximately the same direction, it is implied that they share similar model coefficients up to a scaling factor. These similar models could be clustered for network simplification, as we will discuss in Section~\ref{sec:simplify}. For the BostonHouse example, along around $30^\circ$ (also $150^\circ$, $270^\circ$) direction there are multiple LLMs that are similar to each other. However, for the CoCircles example, the LLMs are seen to scatter around $360$ degrees; see also Figure~\ref{fig: Ex2Profile} (left).

	Those single-instance or single-class regions are resulted from the overparameterized networks, and they are usually redundant. The reliability of the estimated LLMs for these small regions are questionable. Such small regions/LLMs can be either skipped or merged into their neighboring medium/large regions. In Section~\ref{sec:simplify}, we discuss a novel merging strategy for dealing with the small or similar LLMs, which would not only simplify the complex network but also lead to enhanced model interpretability and sometimes improved prediction performance.

	\subsection{Local vs. Global Performance}
\begin{figure}[htp!]
		\cn{\includegraphics[width=0.32\textwidth]{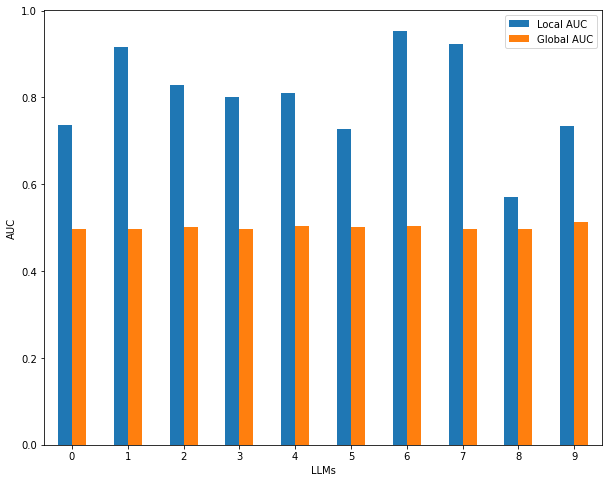}
			\ \ \includegraphics[width=0.32\textwidth]{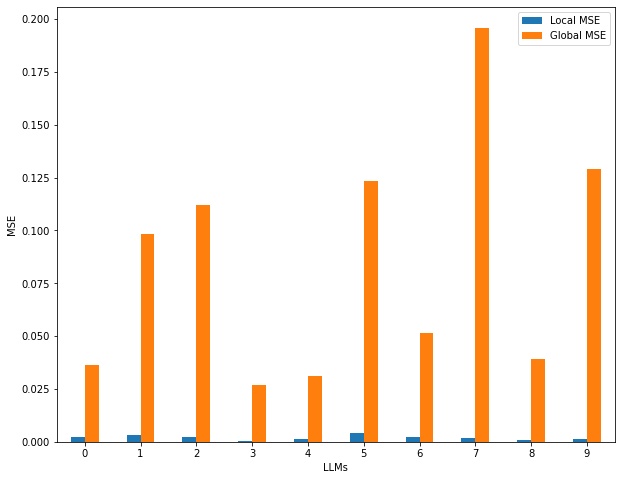}
			\ \ \includegraphics[width=0.32\textwidth]{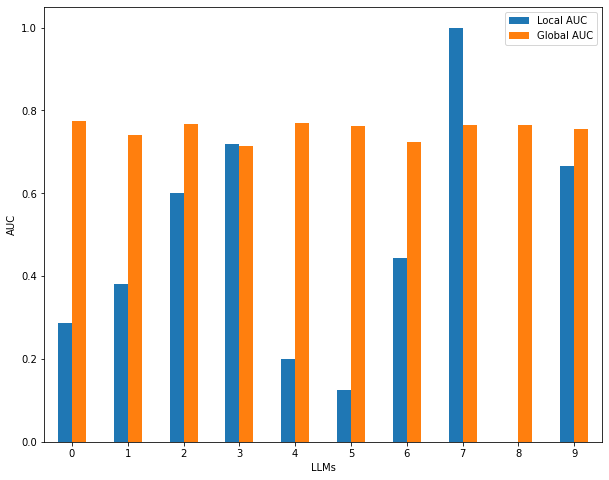}}
		\caption{Diagnostics of extrapolation strengths of top-10 LLMs by comparing their local and global performances. Left: CoCircles (AUC); Center: BostonHouse (MSE); Right: FicoHeloc (AUC)}\label{fig:DiagPerf}
	\end{figure}

	To answer Q2 whether a target LLM is strong or weak, we assess both the local performance within its own region, but also the global performance across other regions.  This latter can be referred to {\em extrapolation strength diagnostics}, which can be done by comparing the local vs. global performances in terms of MSE (regression) or AUC (classification); refer to the last two columns in Figures~\ref{fig:Ex2_Output} and \ref{fig:Ex3_Output}. Alternatively, we can use the side-by-side bar charts for comparing local and global performances of the top-ranked LLMs; see Figure~\ref{fig:DiagPerf} for the CoCircles, BostonHouse, and FicoHeloc examples.

	By comparing local vs. global performance of a target LLM, we have the following rules of extrapolation strength diagnostics: 
	\begin{itemize}
		\item If performance is good locally but poor globally, it is said to have poor extrapolation strength; e.g. almost all the top-ranked LLMs in the CoCircles and BostonHouse examples in Figure~\ref{fig:DiagPerf}. 
		\item If performance is good both locally and globally, it is said to have good extrapolation strength;
		e.g. the ${3, 2, 9}$-indexed LLMs in the FicoHeloc example in Figure~\ref{fig:DiagPerf}. 
		\item If performance is poor locally but good globally, it is said to have extraordinary extrapolation strength; e.g. the ${0,1, 4, 5, 6, 8}$-indexed LLMs in the FicoHeloc example in Figure~\ref{fig:DiagPerf}. 
	\end{itemize}
	Here for the FicoHeloc example, there exist many LLMs with extraordinary extrapolation strength. It can be found these LLMs share similar model coefficients. Furthermore, as we will discuss in the next section, the input-output relationship in FicoHeloc can be approximated well by a regularized logistic model. In this case, although the ReLU net is highly overparameterized, it can successfully identify the underlying LLMs with a sacrifice on local performance but extraordinary global performance. 
	
	\section{LLM-based Network Simplification}\label{sec:simplify}
	
	In this section, two strategies are introduced to simplify the extracted LLMs, including merging and flattening. The former reduces the number of LLMs by merging the locally homogeneous LLMs while the latter further simplifies merged LLMs into a single hidden layer ReLU network. For the merged LLMs, we also discuss how to perform local inference with statistical rigor. 
	
	
	\subsection{Merging}
	Two LLMs can be merged as their corresponding regions are nearby with similar local linear coefficients and intercepts. To satisfy this principle, agglomerative clustering with connectivity constraint is employed to merge LLMs. In specific, each LLM is initialized as a single cluster and different clusters with similar coefficients and intercepts will be merged hierarchically. The similarity between two LLMs is defined as the Euclidean distance between their local linear coefficients and intercepts, subject to the connectivity constraint. The connectivity matrix is calculated by the $T$ nearest neighbors for each LLM, according to their region centers 
	\begin{equation}
	\mathbf{\mu}_{\P} = \frac{1}{|\RR^{\P}_{\rm train}|}\sum_{\x_i \in \RR_{\rm train }^{\P}} \x_i, \quad \text{for } \P \in \PP_{\rm {train }}. \label{llm_centers}
	\end{equation}
	By incorporating the connectivity constraint in agglomerative clustering, the merged LLMs would satisfy both locality and homogeneity.
	
	After merging, it is possible that some clusters only contain a small number of samples. These small clusters are trivial for interpretation. In practice, we use a post-hoc processing step to merge these small clusters to their nearby large clusters. The small and large clusters can be distinguished by a sample size threshold, e.g., $\tau = 30$. We use {\sf AgglomerativeClustering} in {\sf sklearn.cluster} for agglomerative clustering and the connectivity matrix is obtained by {\sf kneighbors\_graph} in {\sf sklearn.neighbors}. The number of clusters $K$ can be tuned using grid search, e.g., from $\{1, 2, 3, 4, 5, 6, 7, 8, 9, 10, 15, 20\}$. The number of neighbors $T$ is initialized to be a small percentage of the total number of LLMs extracted from the training data, e.g. $1\%$ of $\left|\PP_{\rm train}\right|$. If this value is too small to be satisfied, then we increase it until all LLMs can be eventually connected into one cluster. 
	
	Note some new data may not belong to any existing LLMs discovered in the training set, such that the region centers of new LLMs are not available. For simplicity, we use these new samples themselves to represent their corresponding region centers, and the $T$ nearest LLMs can be accordingly detected.
	
\begin{figure*}[htp!]
		\centering
		\includegraphics[width=0.7\textwidth]{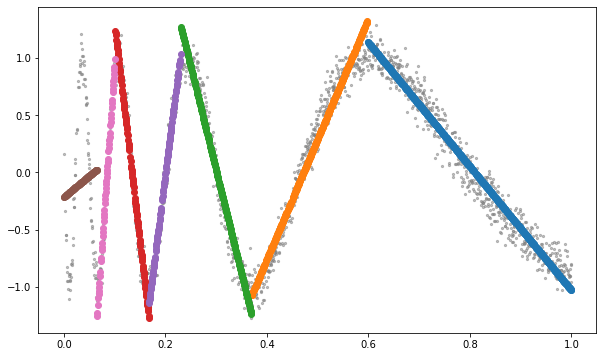}
		\caption{Illustration of ChirpWave after merging.}\label{chirpwave_merge}
		
		\bigskip\bigskip
		
		\subfloat[Merged Classifiers]{
			\label{circle_merge_classifier} 
			\includegraphics[width=0.48\textwidth]{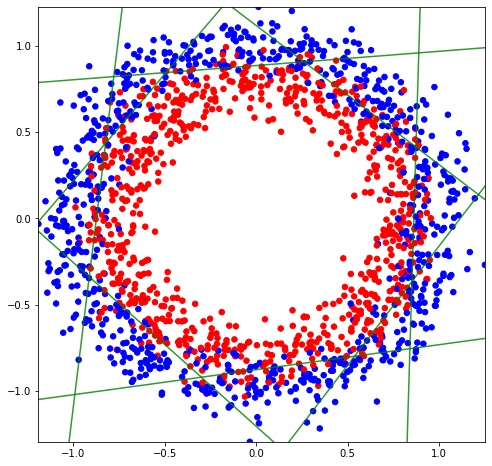}}
		\subfloat[Activation Regions]{
			\label{circile_merge_regions} 
			\includegraphics[width=0.46\textwidth]{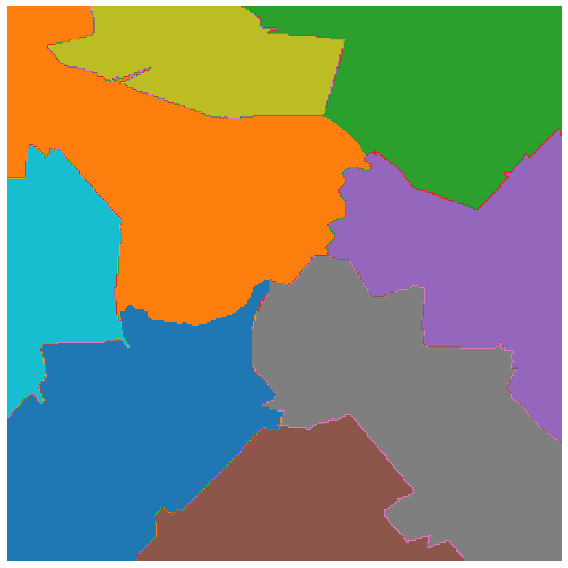}}
		\caption{Illustration of CoCircles after merging.}\label{circle_merge}
\end{figure*}

\begin{figure*}[htp!]
		\centering
		\includegraphics[width=0.7\textwidth]{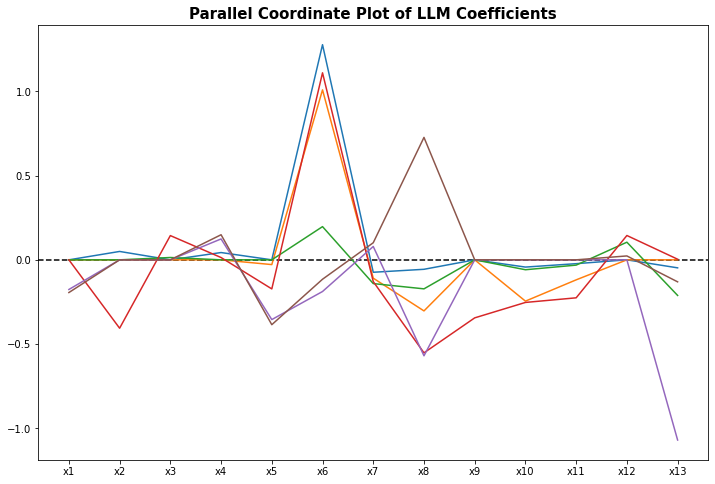}
		\caption{Parallel coordinate plot of the merged network (Data: BostonHouse)}\label{BostonHouseMergePC}
		
		\bigskip		
		\includegraphics[width=0.7\textwidth]{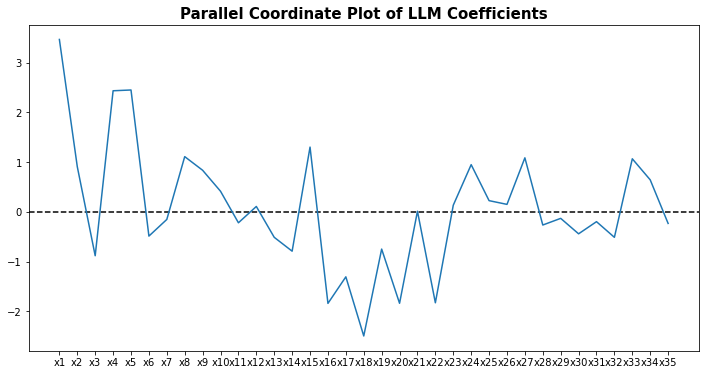}
		\caption{Parallel coordinate plot of the merged network (Data: FicoHeloc)}\label{FICOMergePC}
\end{figure*}

	Finally, we refit a local linear model for each cluster. In specific, two kinds of linear models are considered, i.e., simple GLM for low-dimensional problems or regularized GLM for high-dimensional tasks. The overall merging procedures are summarized in Algorithm~\ref{algo_merge}.
	\begin{itemize}
		\item {\bf Refit model: GLM}. For low-dimensional problems, a simple GLM will be sufficient to capture the local patterns and the resulting model can be largely simplified and is easy to interpret. For the two synthetic datasets, Figures~\ref{chirpwave_merge} and~\ref{circle_merge} present the new partitioning regions and local linear models after merging. It can be observed that a lot of the trivial local regions are merged, and the resulting large regions are close to their corresponding ground truth settings.
		
		\item {\bf Refit model: Regularized GLM}. High-dimensional tasks often suffer from multicollinearity problems and many coefficients can also be hard to interpret. Therefore, we can use the $\ell_{1}$- or $\ell_{2}$-regularized GLM to make the local linear models more stable and interpretable. In Figure~\ref{BostonHouseMergePC}, the parallel coordinate plot of merged LLMs is provided for the BostonHouse dataset. According to this figure, 6 clusters are automatically formed. The corresponding plot for the FicoHeloc dataset is provided in Figure~\ref{FICOMergePC}, in which it is found that only one cluster is left. This means that a deep neural network is over complicated for this dataset, and a logistic regression model is sufficient to capture all the patterns.
	\end{itemize}	
	
	\begin{algorithm}[h!]
		\caption{Merging Algorithm} \label{algo_merge}
		\begin{algorithmic}[1]
			\Require $\{\x_i, y_i\}_{i\in[n]}$, Unwrapped LLMs, $K$ (Number of Clusters), $T$ (Number of Neighbors), $\tau$ (Threshold for Small Clusters).
			\State Calculate region centers $\mathbf{\mu}_{\P}$ of each LLM.
			\State Obtain the $T$ nearest neighbors for each LLM and construct the connectivity matrix.
			\State Cluster $\{(\tilde{\boldsymbol{w}}^{\P}, \tilde{b}^{\P})\}_{ {\P \in \PP_{\rm train }}}$ hierarchically into $K$ groups s.t. connectivity constraint.
			\State Find large clusters (sample size larger than $\tau$) and small clusters (the remaining).
			\State Merge small LLM clusters to their nearest large clusters.
			\State Refit a (regularized) GLM for each new merged LLM.
		\end{algorithmic}
	\end{algorithm}
	
	\subsection{Local Inference}
	The merged LLMs unwrapped from deep ReLU networks can be used for local inference with $p$-values and confidence intervals, which leads to an exact and consistent type of local interpretability. In contrast, the state-of-the-art post-hoc interpretation methods including LIME and SHAP \cite{ribeiro2016should, lundberg2017unified} are non-exact and varying with input perturbations. 
		
	For any input instance $\x\in \XX_{\rm train}$, it is straightforward to determine its activation pattern in a pre-trained $\NN_{\rm train}$; so we can locate the local activation region $\x$ belongs to. The local interpretability of $\x$ is translated to its local region with similar instances. When the sample size within the local region is sufficient, statistical inference can be conducted for the corresponding merged LLMs with the local region support. 
	
	For a given merged local region, denote the estimated local linear coefficients by  $\hat\bb = [\tilde{b}^{\P}, \wtilde^{\P}]$ and the design matrix of $n_1$ local training instances by $\X\in\RRR^{n_1\times d}$. For regression problem, the covariance matrix of $\hat\bb$ can be computed by 
	$\bm\Sigma = \hat\sigma^2\left(\X^T\X\right)^{-1}$, 
in which the variance estimate is given by $\hat\sigma^2 = \|\hat\y - \y \|^2/(n_1 - d-1)$. Then, the Wald test of each individual coefficient $\hat\beta_j$ (including the intercept) uses the $t$-statistic 
$t_j = \hat\beta_j/\sqrt{\bm\Sigma_{jj}}$, which gives  $p$-value and  $95\%$ confidence interval based on the null Student t-distribution with degrees of freedom $n_1 - d - 1$. 
For binary classification, the local inference can be also conducted following the theory of logistic regression. In this case, the covariance matrix of $\hat\bb$ can be computed by 
	$$
	\bm\Sigma = \left(\X^T\W\X\right)^{-1}, \quad\mbox{with } \W = \mbox{diag}\Big(\big\{\hat{p}_i(1-\hat{p}_i)\big\}_{i=1}^{n_1}\Big),
	$$
	where the predicted probability for each instance is given by $(1+\exp\{-\hat\bb^T\x_i\})^{-1}$. Then, the Wald test uses the $z$-statistic $z_j = \hat\beta_j/\sqrt{\bm\Sigma_{jj}}$ with null distribution $N(0,1)$. 
	
\begin{figure}[htp!]
		\cn{\includegraphics[width=0.4\textwidth]{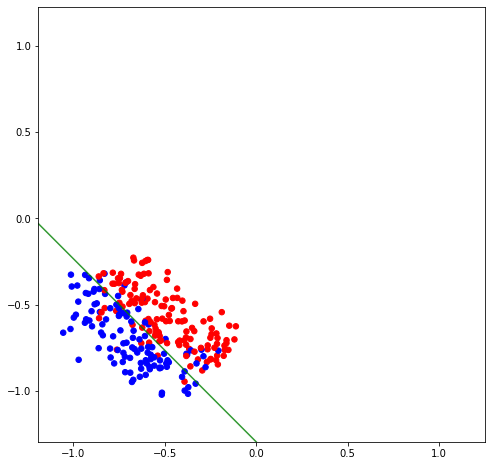}
			\ \ \ \includegraphics[width=0.4\textwidth]{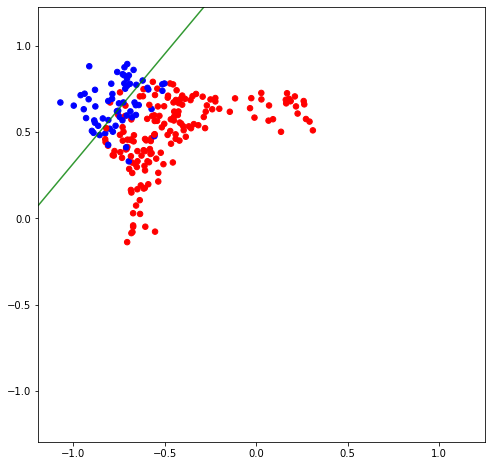}}
		\caption{Top two regions of merged LLMs with training instances (Data: CoCircles).}\label{fig: Ex2LocInf}
	\end{figure}

	For a quick demonstration of local inference, consider the top two merged LLMs for the CoCircles example, as visualized in Figure~\ref{fig: Ex2LocInf}. For each region, we conducted the local inference procedure with testing results shown in Table~\ref{tab:Ex2LocInf}. The intercept and coefficients are tested all significant (with level $0.05$) for both regions. These testing results are reasonable upon a visual check as shown in Figure~\ref{fig: Ex2LocInf}. 	

	\begin{table}[hbtp!]	
		\centering
		\renewcommand\tabcolsep{8pt}
		\renewcommand\arraystretch{1.2}
		\caption{Local inference results for top two merged regions (Data: CoCircles).}	\label{tab:Ex2LocInf}
		{\sf\begin{tabular}{l|rrrrrr}
				\hline\hline
				\vspace{-0.1cm}
				{\bf Region 1} &    coef & std\_err &       z & p-value &  [0.025 &  0.975] \\ \hline
				intercept                          &  5.3201 &   0.9076 &  5.8616 &     0.0 &  3.5412 &  7.0990 \\
				x1                                 &  4.3638 &   0.7963 &  5.4802 &     0.0 &  2.8031 &  5.9245 \\
				x2                                 &  4.1135 &   0.8906 &  4.6188 &     0.0 &  2.3679 &  5.8590 \\ \hline\hline
				{\bf Region 2}                     &    coef & std\_err &       z & p-value &  [0.025 &  0.975] \\ \hline
				intercept                          &  5.5461 &   0.8524 &  6.5064 &  0.0000 &  3.8754 &  7.2168 \\
				x1                                 &  4.4345 &   0.9225 &  4.8069 &  0.0000 &  2.6264 &  6.2426 \\
				x2                                 & -3.4950 &   0.9619 & -3.6335 &  0.0003 & -5.3802 & -1.6098 \\ \hline\hline
		\end{tabular}}		
	\end{table}
	
\begin{figure}[htp!]
		\centering
		\includegraphics[width=0.8\textwidth]{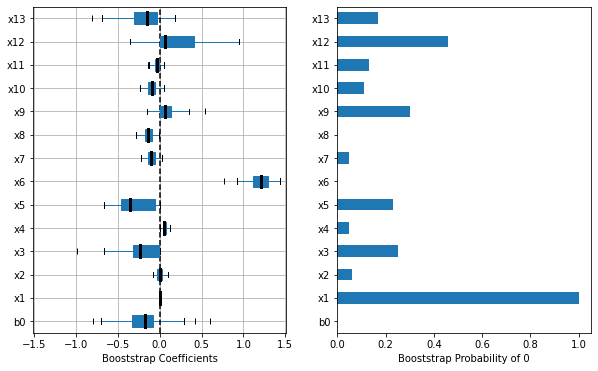}
		\caption{Bootstrap inference for BostonHouse dataset.}\label{Ex3_MergeInference}
	\end{figure}

	Besides, local inference can be performed for $\ell_{1}$-regularized local linear models. For demonstration, the left sub-figure in Figure~\ref{Ex3_MergeInference} shows the coefficients and intercepts for the BostonHouse dataset based on 100 bootstrap replicates; while the right sub-figure presents the probability that the coefficient or intercept is reduced to zero. For instance, {\sf x6} is tested to be a significantly positive feature of the response. 
	
	
	
	\subsection{Flattening}
	As LLMs are merged into a small number of clusters, we may do further simplification by flattening them into a single hidden layer ReLU network. That is, a new single hidden layer ReLU network is created where the number of hidden neurons is set to the number of merged LLM clusters. The hidden layer weights and biases are configured by the local linear coefficients and intercepts of the merged LLM clusters. The output layer weights and bias can be estimated via GLM. After that, we fine-tune the flattened network for some epochs, to achieve better predictive performance. In the original deep ReLU network, each LLM corresponds to a linear model that is only active locally. But as the network is flattened, each of the original local regions is now affected by an ensemble of multiple linear models. Below, we present Algorithm~\ref{algo_flatten} for the flattening strategy.
	\begin{algorithm}[h]
		\caption{Flattening Algorithm} \label{algo_flatten}
		\begin{algorithmic}[1]
			\Require $\{\x_{i}, y_{i}\}_{i\in[n]}$, Merged LLM Clusters.
			\State Collect local linear coefficients and intercepts from each merged LLM cluster.
			\State Configure the first hidden layer weights and biases by the extracted local linear coefficients and intercepts, respectively.
			\State Forward propagate the data and configure the output layer by GLM.
			\State Fine-tune the flattened network using SGD-based optimizers. 
		\end{algorithmic}
	\end{algorithm}

	\begin{table}[htp!]
		\centering
		\renewcommand\tabcolsep{10pt}
		\renewcommand\arraystretch{1.2}
		\caption{Average results of ReLU Net, Merged Net, Flattened Net and SLFN, based on 100 Monte Carlo replicates (with standard deviations in parentheses). The last column indicates the average number of clusters used in Merge-Net and FL-Net, as well as the number of hidden nodes in SLFN.} \label{merge_comparison}
		\begin{tabular}{c|cccc|c}
			\hline\hline
			Data                       & ReLU-Net & Merge-Net &  FL-Net  &   SLFN   & n\_cluster \\ \hline
			\multirow{2}{*}{\shortstack{ChirpWave\\(MSE)}}  &  0.0486  &  0.0721   &  0.0839  &  0.3351  &  10.2800   \\
			& (0.0356) & (0.0540)  & (0.0323) & (0.1021) &  (2.4498)  \\ \hline
			\multirow{2}{*}{\shortstack{CoCircles\\(AUC)}}  &  0.9174  &  0.8720   &  0.8845  &  0.8280  &   9.3800   \\
			& (0.0157) & (0.0288)  & (0.0296) & (0.1387) &  (2.9691)  \\ \hline
			\multirow{2}{*}{\shortstack{BostonHouse\\(MSE)}} &  0.0074  &  0.0079   &  0.0087  &  0.0132  &   3.9500   \\
			& (0.0027) & (0.0021)  & (0.0031) & (0.0087) &  (1.2913)  \\ \hline
			\multirow{2}{*}{\shortstack{FicoHeloc\\(AUC)}}  &  0.8002  &  0.8050   &  0.7962  &  0.7545  &   1.5500   \\
			& (0.0091) & (0.0088)  & (0.0107) & (0.1102) &  (1.0897)  \\ \hline\hline
		\end{tabular}
	\end{table}	

	In Table~\ref{merge_comparison}, we compare the test set predictive performance of the original deep ReLU network (ReLU-Net), merged LLMs (Merge-Net), flattened network (FL-Net), and single hidden layer feed-forward network (SLFN). Note that SLFN is configured to have the same number of hidden nodes as is used in FL-Net. The only difference between FL-Net and SLFN lies in that they have different initialization methods. That is, the hidden layer of FL-Net is initialized by the refitted local linear models of Merge-Net, while SLFN's hidden layer is randomly initialized. Compared with the original ReLU-Net, the proposed Merging strategy not only reduces the model complexity but also may improve the generalization performance, e.g., on the FICO dataset. For the other datasets, its test performance is still close to that of ReLU-Net. The flattening strategy can further reduce model complexity. Given the same model complexity, FL-Net has significantly superior predictive performance as compared to SLFN. This indicates the local linear coefficients of Merge-Net make a good initialization for FL-Net.  

	\section{Case Study on Home Lending Dataset}\label{sec-homelending}
	In this section, we present a real-world case study of predicting credit default for home lending. The purpose of the case study is to illustrate how the methodology described in this paper is applied in real-life environments. In a regulated financial institution, model interpretability is a requirement for the purpose of evaluating the conceptual soundness of models as well as to explain to model users or customers the decision made by models. Part of the conceptual soundness requirement, the effects of variables must be consistent with the expected business knowledge. For example, lower credit quality (e.g. FICO score) drives a higher probability of default. The lack of ability to evaluate the conceptual soundness of DNNs has been a major obstacle for credit risk applications. In this case study, we demonstrate that by extracting the equivalent local linear models for ReLU-Net we can provide a strong mechanism to evaluate model conceptual soundness equivalent to the traditional statistical linear regression models. The model diagnostics are applied to identify local linear models that can be improved. Further, model simplification is applied to enhance both model interpretability and model performance. It will be shown below how the large number of local linear models from ReLU-Net are reduced into a smaller number of local linear models and at the same time improve the performance from the original networks.
	
	We modified the dataset in this case study for confidentiality purposes without affecting the ability to demonstrate the methodology. The response variable of interest is binary with the label 0 as ``customer does not default'' and 1 as ``customer defaults'' based on various loan characteristics as well as other auxiliary variables totaling 55 predictors where the top 19 are summarized in Table~\ref{Homelending_Introduction}. The dataset has 1 million instances, of which about 1\% are defaulting cases. We sampled and split the data into training and testing sets, each with approximately $\frac{1}{3}$ of the total instances, using the same split as explained elsewhere. Furthermore, we make the dataset balanced, by randomly selecting the data from label 0 class to make sure each class has equal distribution in the final dataset. 
	
	\begin{table}[!h]
		\begin{center}
			\caption{Important features and their definition in the Homelending 
				Dataset.} \label{Homelending_Introduction}
			\begin{tabular}{ll}
				\hline\hline
				Variable    &                                Meaning                                 \\ \hline
				fico0          &                                FICO at prediction time                                 \\ 
				ltv{\_}fcast      &                             Loan to value ratio forecasted                             \\ 
				sato2          &                             Spread at time of origination                              \\ 
				unemprt         &                                   Unemployment rate                                    \\ 
				dlq{\_}new{\_}delq0   &                Delinquency status: 1 means current, 0 means delinquent                 \\ 
				orig{\_}hpi       &                             Origination house price index                              \\ 
				ifix15         &                             Indicator for Fixed 15yr Loan                              \\ 
				h            &                                   Prediction horizon                                   \\ 
				premod{\_}ind      &                Time indicator: before 2007Q2 financial crisis vs. after                \\ 
				qspread         &                           Spread: note rate -- mortgage rate                           \\ 
				iarm          &                                 Indicator for ARM Loan                                 \\ 
				totpersincyy      &                      Total personal income year over year growth                       \\ 
				orig{\_}fico      &                                  FICO at origination                                   \\ 
				hpiyy          &                                        HPI YoY                                         \\ 
				rgdpqq         &                                      Real GDP QoQ                                      \\ 
				orig{\_}cltv      &                                Origination combined ltv                                \\ 
				ifulldoc:        &                            Ifulldoc: Indicator for Full Doc                            \\ 
				balloon{\_}in      &                                 Balloon loan indicator                                 \\ 
				dum{\_}loanpurp{\_}P  & dummy variable of loanpurp=='P'\\ 
				interest{\_}only{\_}in &                              Indicator for Interest Only                               \\ 
				isf           &                                     Single Family                                      \\ \hline\hline
			\end{tabular}
		\end{center}
	\end{table}
	
	\subsection{Model Details}
	For illustration purposes, we use a feed-forward 2 layer neural with ReLU as	the nonlinear activation with 5 neurons in each layer. The model has a performance on the test set of 75.77\% accuracy and ROC AUC of 84.29\% that are comparable to the performance of best models reported earlier~\cite{chen2020adaptive}. Table~\ref{Homelending_llms} shows the summary of the largest regions such as the number of samples and the performance of local linear models from ReLU-Net. There are a total of 62 regions of which 14 (23\%) are single sample instances or a single class.
	
	\begin{table}[!h]
		\begin{center}
			\caption{Summary of 15 Largest Regions Unwrapped from ReLU-Net.} \label{Homelending_llms}
			\begin{tabular}{cccccc}
				\hline\hline
				& Count & Response Mean & Response Std & Local AUC & Global AUC \\ \hline
				0  & 2303 &    0.461     &    0.498     &  0.791   &   0.834    \\
				1  & 615  &    0.668     &    0.471     &  0.766   &   0.824    \\
				2  & 527  &    0.557     &    0.497     &  0.718   &   0.832    \\
				3  & 478  &    0.495     &    0.499     &  0.699   &   0.831    \\
				4  & 320  &    0.043     &    0.204     &  0.641   &   0.829    \\
				5  & 258  &    0.732     &    0.442     &  0.674   &   0.806    \\
				6  & 218  &    0.825     &    0.379     &  0.628   &   0.768    \\
				7  & 206  &    0.038     &    0.193     &  0.761   &   0.784    \\
				8  & 187  &    0.658     &    0.474     &  0.688   &   0.804    \\
				9  & 155  &    0.935     &    0.245     &  0.579   &   0.823    \\
				10 & 138  &    0.217     &    0.412     &  0.631   &   0.820    \\
				11 & 131  &    0.145     &    0.352     &  0.679   &   0.791    \\
				12 & 121  &    0.793     &    0.404     &  0.775   &   0.796    \\
				13 & 116  &    0.129     &    0.335     &  0.606   &   0.815    \\
				14 & 105  &    0.647     &    0.477     &  0.714   &   0.812    \\
				15 & 103  &    0.106     &    0.308     &  0.590   &   0.761    \\ \hline\hline
			\end{tabular}
		\end{center}
	\end{table}
	
\begin{figure}[!htp]
\centering
\includegraphics[width=1.0\textwidth]{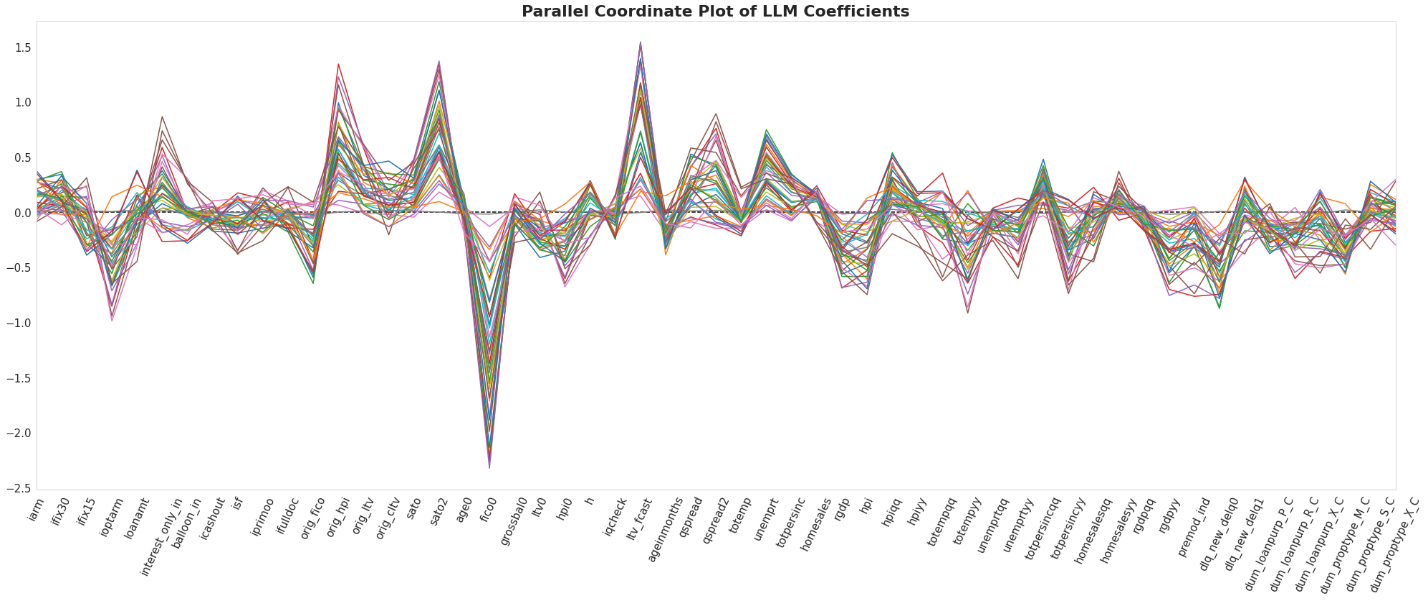}
\caption{PC plot for the Homelending data.}\label{homelending_PC}
\bigskip	
\centering
\includegraphics[width=0.8\textwidth, height=0.7\textwidth]{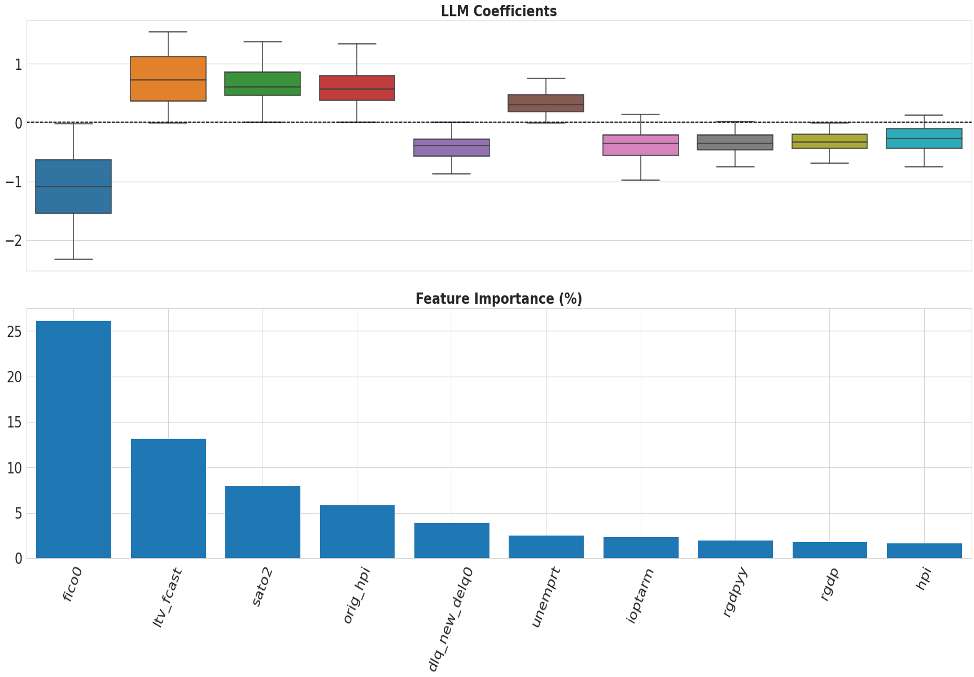}
\caption{Homelending case study: distribution of LLM coefficients (above); most important features (below).} \label{homelending_features_importance}
\end{figure}

	Excluding the single instances and class, the coefficients of local linear 
	models are shown as a parallel coordinate plot in Figure~\ref{homelending_PC}. The top 10 most important variables and the distribution of their coefficients values are shown in Figure~\ref{homelending_features_importance}. 
	
\begin{figure}[!t]
		\begin{center}
			\subfloat[]{
				\includegraphics[width=0.48\textwidth]{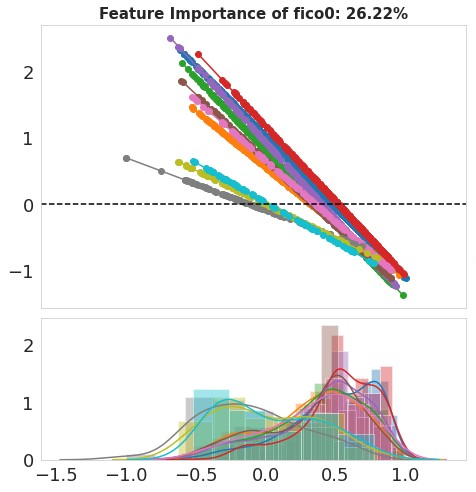}}
			\subfloat[]{
				\includegraphics[width=0.48\textwidth]{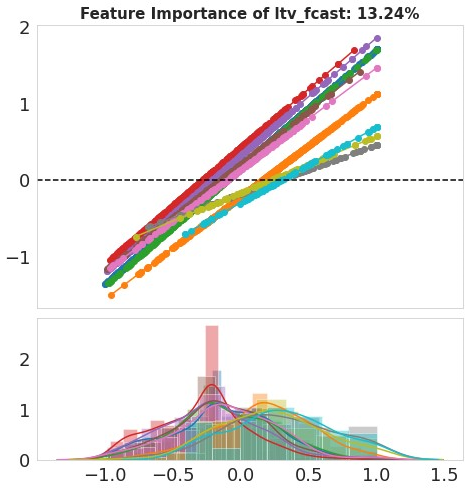}}
			\caption{Profile of two most important variables. (a) FICO and (b) Forecasted Loan to Value (LTV).}
			\label{homelending_Profile}
		\end{center}
	\end{figure}

	The profile of the two most important variables, i.e., FICO and Forecasted Loan to Value (LTV) are shown in Figure~\ref{homelending_Profile}. The higher the FICO the less probability of default while the higher the LTV the higher probability of default, consistent with business intuition.
	
	Notice in Table~\ref{Homelending_llms} there are regions where the response means are either closer to 0 (non-default), 1 (default), or 0.5 (mix of default and non-default) with varying AUCs locally. To further enhance the model, we apply region merging particularly to address less reliable (small) regions where the sample size is small, particularly those regions with single sample size. Here we apply the merging algorithm discussed in~\ref{sec:simplify}. In this section, we present the results from the merging algorithm as mentioned in Section~\ref{sec:simplify}. The merging step reduced the ReLU-Net regions into three regions as shown in Table~\ref{homelending_merged}.
	
	The total AUCs of the merged regions (training AUC= 0.8532, testing AUC = 0.8388), in fact, are better than the original ReLU-Net (training AUC=0.8476, testing AUC=0.8316). This merging step shows that most of the regions created by ReLU-Net are redundant and can be replaced by three LLM that gives better predictive performance. Here, we applied penalized regressions ({\sf sklearn.linear\_model.LogisticRegression} with regularization parameter $C=0.1$) for each merged region. 
	
\begin{figure}[!htp]
\centering
			\includegraphics[width=6.50in,height=2.60in]{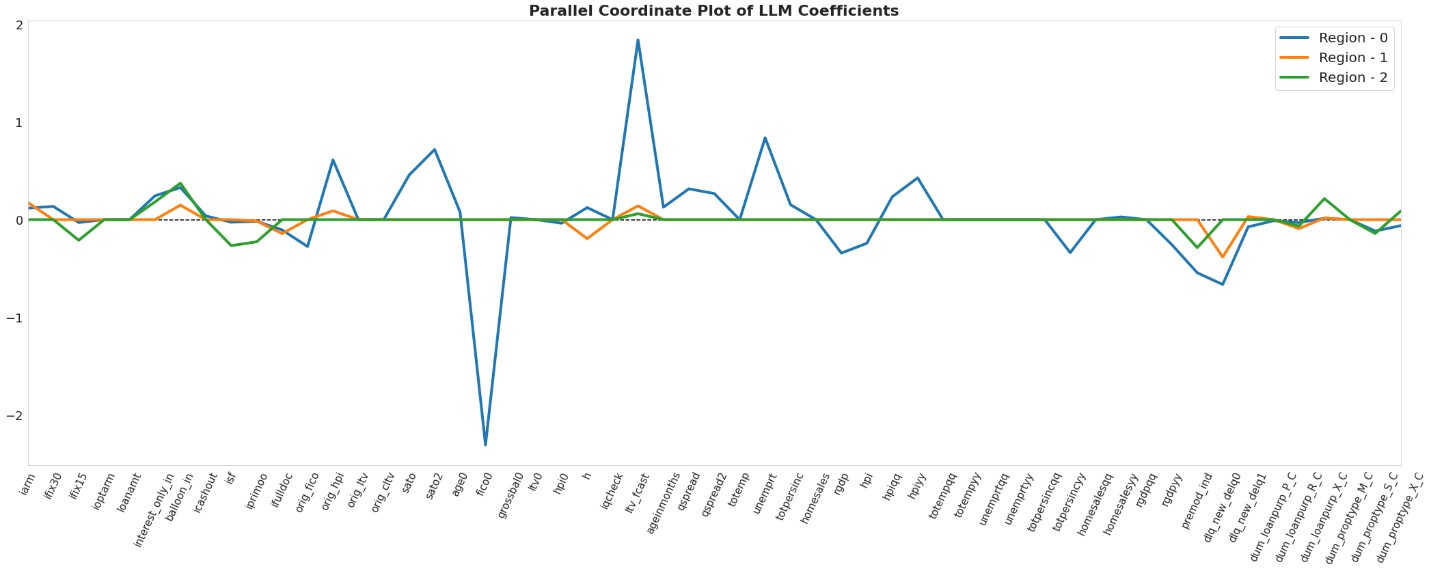}
			\caption{PC plots for the merged regions.}\label{homelending_merged_PC}

\bigskip\bigskip

\cn{\includegraphics[width=0.48\textwidth]{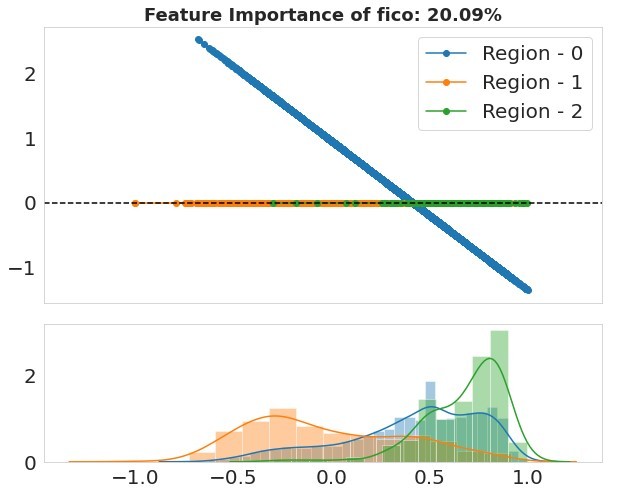}
\ \ \ \includegraphics[width=0.48\textwidth]{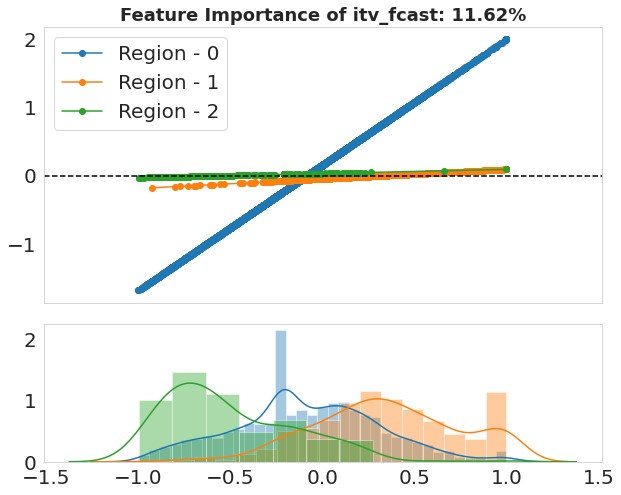}}
\caption{Profile plots of the most important features.}  \label{homelending_merged_Profile}
\end{figure}

	Notice the results of the merging step created three distinct data segmentation (LLMs) with distinct characteristics. Region 0, the largest region with 85.2\% of the data contains mixed of default and non-default cases (response mean near 0.5). Region 1 consists of the majority default cases (90\%) where Region 2 consists of the majority non-default cases (85.4\%). The comparison between local and global AUCs also indicates the distinction among three regions as the local AUCs are better than the global AUCs. It is interesting to observe the parallel coordinate plot of the three LLMs and the corresponding profile of the most important variables shown in Figures~\ref{homelending_merged_PC} and~\ref{homelending_merged_Profile}, respectively. The profile plots indicate three distinct borrower characteristics: (1) High credit quality (high FICO, low LTV), (2) Middle credit quality (mid FICO and LTV), and (3) Low credit quality (low FICO, high LTV). The parallel coordinate plot provides contrast in terms of significant variables that drive the default event. FICO and LTV are important factors for the majority of the loans (Region 0). Delinquency status is the most important variable for the low credit quality (Region 1) where loans that are not currently delinquent will have a lower probability of default. Premod{\_}ind which is the indicator of loan origination before or after the financial crisis is important for the higher credit quality (Region 2). Loans originated post-financial crisis are better performing loans as they are subjected to a higher origination standard. It is also interesting to point out the effect of the ``h'' variable which is the time horizon of prediction, an equivalent concept to baseline hazard rate in the survival analysis model. For lower credit quality, the longer time horizon that the loans do not default the less likely the loans will default (i.e., decreasing hazard rate) while the high-quality loan is the opposite which is consistent with business intuition and credit theory from a stochastic process point of view.

	\begin{table}[!t]
		\begin{center}
			\caption{Summary of Three Merged Regions.} \label{homelending_merged}
			\begin{tabular}{cccccc}
				\hline\hline
				Region & Count & Response Mean & Response Std & Local AUC & Global AUC \\ \hline
				0      & 5602 & 0.460        & 0.498        & 0.831    & 0.854      \\ 
				1      & 747  & 0.904        & 0.293        & 0.680    & 0.584      \\ 
				2      & 219  & 0.146        & 0.353        & 0.987    & 0.736      \\ \hline\hline
			\end{tabular}
		\end{center}
	\end{table}

In the merging process above we retain the original networks for data partitioning and the final three LLMs are applied through a lookup table to map the regions from the original ReLU-Net to the merged regions. If a smaller network is preferred instead of using the original network, the final LLMs can be used directly as an FL-Net discussed in Section~\ref{sec:simplify}. Here, we constructed a simple single hidden layer network with three nodes to represent the LLMs. The result comparison for the FL-Net is summarized in Table~\ref{homelending_Performance}. Note that the FL-Net has comparable performance as the Merged-Net and is better than the original ReLU-Net.
	
	\begin{table}[!t]
	\centering
		\renewcommand\tabcolsep{12pt}
		\renewcommand\arraystretch{1.2}
			\caption{AUC Performance Comparison of ReLU-Net, Merge-Net and FL-Net.} 	\label{homelending_Performance}
			\begin{tabular}{cccc}
				\hline\hline
				& ReLU-Net & Merge-Net & FL-Net \\ \hline
				Training & 0.8476   & 0.8532 & 0.8538    \\ 
				Testing  & 0.8316   & 0.8388 & 0.8368    \\ \hline\hline
			\end{tabular}
	\end{table}

	\section{Conclusion}\label{sec-conclusion}
	We have introduced a novel  \aletheia toolkit for unwrapping the black box of deep ReLU networks.  Unlike post-hoc explainability methods based on crude assumptions or input permutations, the proposed unwrapper is a rigorous approach to intrinsic interpretability of ReLU DNNs, and it enables us to perform direct interpretation, diagnostics, and simplification based on the unwrapped set of local linear models. Multiple examples used in the paper,  including a real case study in home lending credit risk analytics, have demonstrated the novelty and effectiveness of the proposed methods. 
	
	\bibliographystyle{abbrv}
	\bibliography{mybibfile}
	
\end{document}